\documentclass{article}

\usepackage{colm2025_conference}
\microtypesetup{expansion=false}

\usepackage{amsmath,amssymb,amsfonts}
\usepackage{fontspec}
\usepackage{graphicx}
\usepackage{subcaption}
\usepackage{booktabs}
\usepackage{multirow}
\usepackage{algorithm}
\usepackage{algorithmic}
\usepackage[most]{tcolorbox}
\usepackage{tikz}
\usetikzlibrary{shapes.geometric, arrows.meta, positioning, fit, backgrounds, calc, decorations.pathreplacing}
\usepackage{enumitem}
\usepackage{verbatim}
\usepackage{float}
\setlist[enumerate]{leftmargin=*, labelsep=0.5em}
\setlist[itemize]{leftmargin=*, labelsep=0.5em}

\newtcbox{\rolebox}[1][gray]{%
  on line, arc=2pt, boxrule=0.4pt, boxsep=0pt,
  left=2.5pt, right=2.5pt, top=0.6pt, bottom=0.6pt,
  colback=#1!12, colframe=#1!55!black, coltext=#1!42!black,
  fontupper=\sffamily\fontsize{9.1}{8.91}\selectfont\bfseries}
\newcommand{\roleInTable}[1]{%
  \ifcsname rolecol@#1\endcsname
    \rolebox[\csname rolecol@#1\endcsname]{\makebox[2em][c]{\strut#1}}%
  \else
    \rolebox{\makebox[2em][c]{\strut#1}}%
  \fi}
\newcommand{\roleMain}[1]{%
  \ifcsname rolecol@#1\endcsname
    \rolebox[\csname rolecol@#1\endcsname]{#1}%
  \else
    \rolebox{#1}%
  \fi}

\expandafter\def\csname rolecol@ww\endcsname{red}
\expandafter\def\csname rolecol@vl\endcsname{blue}
\expandafter\def\csname rolecol@sr\endcsname{teal}
\expandafter\def\csname rolecol@wt\endcsname{violet}
\expandafter\def\csname rolecol@ht\endcsname{orange}
\expandafter\def\csname rolecol@ww-1\endcsname{red}
\expandafter\def\csname rolecol@ww-2\endcsname{red}
\expandafter\def\csname rolecol@ww-3\endcsname{red}

\definecolor{ajBlue}{HTML}{E8F1FF}
\definecolor{ajBlueLine}{HTML}{4D79B8}
\definecolor{ajGreen}{HTML}{E9F7EF}
\definecolor{ajGreenLine}{HTML}{3D8B62}
\definecolor{ajOrange}{HTML}{FFF1DB}
\definecolor{ajOrangeLine}{HTML}{C67A1B}
\definecolor{ajPurple}{HTML}{F2ECFF}
\definecolor{ajPurpleLine}{HTML}{7A60B8}
\definecolor{ajGray}{HTML}{F5F6F8}
\tikzset{
  ajBox/.style={rectangle, rounded corners=3pt, draw=black!55, line width=0.5pt, fill=white, align=center, font=\small, inner sep=4pt, minimum height=0.72cm},
  ajSmall/.style={ajBox, font=\scriptsize, inner sep=3pt, minimum height=0.54cm},
  ajClient/.style={ajBox, fill=ajBlue, draw=ajBlueLine},
  ajServer/.style={ajBox, fill=ajGreen, draw=ajGreenLine},
  ajControl/.style={ajBox, fill=ajOrange, draw=ajOrangeLine},
  ajFeedback/.style={ajBox, fill=ajPurple, draw=ajPurpleLine},
  ajStore/.style={ajBox, fill=ajGray, draw=black!45},
  ajGroup/.style={rectangle, rounded corners=5pt, draw=black!35, line width=0.5pt, dashed, fill=black!1, inner sep=6pt},
  ajArrow/.style={-{Stealth[length=2.5mm]}, line width=0.75pt, draw=black!55},
  ajDataArrow/.style={-{Stealth[length=2.5mm]}, line width=0.9pt, draw=ajBlueLine},
  ajTrainArrow/.style={-{Stealth[length=2.5mm]}, line width=0.9pt, draw=ajGreenLine},
  ajControlArrow/.style={-{Stealth[length=2.5mm]}, line width=0.9pt, draw=ajOrangeLine},
  ajFeedbackArrow/.style={-{Stealth[length=2.5mm]}, line width=0.9pt, draw=ajPurpleLine, dashed},
}

\title{AgentJet: A Distributed Swarm Training Framework for Agentic Reinforcement Learning}

\author{
  Qingxu Fu,
  Boyin Liu,
  Shuchang Tao,
  Zhaoyang Liu,
  Cheng Chen,
  Xuanfa Jin,
  Rong Zhu,
  Bolin Ding \\
  [1em]
  Tongyi Lab, Alibaba Group \thanks{This work is conducted at Tongyi Lab, Alibaba Group. Correspondence: \texttt{\{fuqingxu.fqx, liuboyin.lby, taoshuchang.tsc, liuzhaoyang.lzy, chencheng.cc, jinxuanfa.jxf, zhurong.zr, dingbolin.dbl\}@alibaba-inc.com}} \\
}

\begin{document}

\maketitle

\begin{abstract}
Training reinforcement learning (RL) policies for large language model (LLM) agents requires optimizing multi-turn execution trajectories that interact with external environments.
These trajectories exhibit runtime fragility, single-model constraints, runtime incompatibility across tasks, and redundant context that general-purpose RL training frameworks were not designed to handle.
We present \textbf{AgentJet}, a distributed swarm training framework that addresses these challenges through a decoupled multi-node architecture.
AgentJet treats the server--client topology as a configurable part of the training system: different connections among trainable swarm servers and isolated swarm clients instantiate different training regimes without changing the trainer implementation.
\textbf{Swarm server} nodes host trainable models and run optimization on GPU clusters, whereas \textbf{swarm client} nodes execute arbitrary agents on arbitrary devices and communicate with the servers through standard, OpenAI-compatible serving APIs.
By reconfiguring this topology, the same serving interface supports multiple training paradigms that are difficult to realize in centralized frameworks:
(1)~Heterogeneous multi-model RL, which trains multi-agent teams whose members are driven by distinct LLMs;
(2)~Mixed-task training with isolated and detachable agent runtimes;
(3)~Fault-tolerant execution that prevents external environment failures from interrupting the training process;
and (4)~Live code iteration, which enables a Read--Eval--Print--Loop (REPL) style workflow that edits agents against a live training run by hot-swapping swarm client nodes.
To support efficient RL in multi-model, multi-task, and multi-agent settings, AgentJet introduces a context-tracking module with timeline merging, which consolidates redundant context and reduces actor-update time by 6.25$\times$ on the AppWorld benchmark.
Built on the same detachable-client substrate, AgentJet further provides an automated research system that takes a research topic as input and conducts long-horizon, multi-day RL studies on large-scale clusters with reduced human intervention.
AgentJet is fully open-source and compatible with agents built on any agent harness system that issues standard LLM inference requests.
\end{abstract}

\section{Introduction}
\label{sec:introduction}

Large language models (LLMs) have evolved from static text generators into autonomous agents capable of multi-turn reasoning, tool invocation~\citep{yao2023react,schick2023toolformer}, and complex workflow execution~\citep{openai2024o1,deepseek2025r1,qwen3}. Reinforcement learning (RL) has emerged as a central paradigm for developing these capabilities: DeepSeek-R1~\citep{deepseek2025r1} demonstrated that advanced reasoning behaviors, including self-reflection, verification, and dynamic strategy adaptation, can emerge through RL without supervised fine-tuning, while the OpenAI o1 series pioneered inference-time scaling through chain-of-thought reasoning~\citep{wei2022cot} trained with large-scale RL~\citep{openai2024o1}. As models transition from single-turn reasoning to long-horizon agentic tasks such as coding across entire repositories~\citep{jimenez2024swebench}, navigating web browsers~\citep{zhou2024webarena,deng2023mind2web}, and orchestrating office workflows, the demands placed on RL training infrastructure have grown substantially. In this shift, the unit of RL optimization moves from a single prompt-response pair to an entire execution trajectory: a single training sample now spans multiple inferences, environment mutations, tool failures, and delayed rewards, so that the object being optimized is no longer a model completion but an executing agent system.

A rich ecosystem of RL training frameworks has emerged to meet this challenge. General-purpose RLHF and post-training frameworks such as OpenRLHF~\citep{hu2024openrlhf}, veRL (HybridFlow)~\citep{sheng2024hybridflow}, and slime~\citep{thudm2025slime} have established efficient distributed architectures for post-training LLMs. OpenRLHF pioneered a Ray~\citep{moritz2018ray} + vLLM~\citep{kwon2023vllm} distributed architecture that orchestrates actor, reward, reference, and critic models across GPU clusters with hybrid engine scheduling. veRL introduced a hybrid single-controller/multi-controller programming model that decouples RL control flow from computation flow, achieving state-of-the-art throughput across PPO~\citep{schulman2017ppo}, GRPO~\citep{shao2024deepseekmath}, and other algorithms, while slime connects Megatron-LM training with SGLang-based rollout generation for RL scaling.

However, training \textit{agentic} models, in which the LLM must interact with external environments over multiple turns, invoke tools, manage context, and coordinate with other agents, introduces qualitatively different challenges that general-purpose RLHF frameworks were not designed to address. Agentic RL trajectories are orders of magnitude longer than single-turn reasoning chains, involve heterogeneous external environments such as code sandboxes, browsers, and APIs, and exhibit high variance in completion times. These demands have catalyzed a new wave of agent-native RL frameworks. Forge by MiniMax~\citep{minimax2026forge} introduces a middleware abstraction layer that fully decouples the training-inference engine from the agent scaffolding, treating agents as black boxes. AReaL~\citep{fu2025areal} pioneered fully asynchronous training-inference decoupling, achieving 3--5$\times$ speedup in agent search scenarios. Agent Lightning~\citep{luo2025agentlightning} converts agent experience into state-action-reward transitions that can be consumed by any RL algorithm. OpenTinker~\citep{zhu2026opentinker} advances the separation-of-concerns paradigm with a Reinforcement Learning as a Service (RLaaS) architecture. OpenClaw-RL~\citep{wang2026openclawrl} explores every agent interaction as a live training signal for continuous online policy optimization.

Despite this rapid progress, existing frameworks share several critical limitations when applied to real-world LLM agent training:

\begin{itemize}
\item \textbf{Multi-Model Constraints:} Most current systems optimize a single policy at a time, whereas real-world agentic applications increasingly field heterogeneous teams in which different roles are backed by models of different sizes (e.g., a 7B executor alongside a 32B planner), each requiring its own trainable policy.

\item \textbf{Runtime Fragility:} Most frameworks co-locate agent rollout environments and model training in the same cluster. When an agent depends on heavyweight components such as Docker containers, browser automation, or database connections, a single environment failure can crash the entire pipeline, discard all progress since the last checkpoint, and even cascade across the many concurrent episodes that share the same process.

\item \textbf{Iteration Overhead:} Editing agent code or reward functions forces a full restart of the training process, reloading the model, reinitializing the inference engine, and discarding any in-flight episodes, so that even a trivial change, such as a reward coefficient change, costs 5--10 minutes.

\item \textbf{Redundant Context:} As an agent interacts with its environment over many turns, its conversation accumulates substantial redundancy (repeated system prompts, tool definitions, and observation histories), which the optimizer reprocesses on every policy-gradient update.

\item \textbf{Runtime Incompatibility:} Mixed-task training jointly optimizes one policy over tasks with mutually incompatible runtimes, so it demands strong isolation between task environments, which monolithic trainers cannot provide.
\end{itemize}

In this paper, we introduce \textbf{AgentJet}, a distributed framework that establishes a unified \textbf{serving-optimization protocol} in place of tightly coupled rollout loops, structurally decoupling the training pipeline into two node types. \emph{Swarm servers} manage trainable models, inference engines, optimizers, and sample pools on GPU clusters; \emph{swarm clients} execute arbitrary agent loops, environments, and evaluators across heterogeneous devices, interacting with servers through standard OpenAI Chat/Responses-compatible API requests augmented with temporary per-episode routing credentials. This routing associates each inference request with the correct training episode, target model, and reward signal while leaving client-side agent logic largely unmodified. By elevating the LLM serving interface to a formal \textbf{execution-optimization abstraction}, AgentJet supports four properties required by agentic RL: \emph{framework compatibility} (agents built with LangChain~\citep{langchain}, AgentScope~\citep{gao2024agentscope}, the OpenAI Agents SDK~\citep{openaiagents2026}, or raw HTTP clients), \emph{runtime isolation} (incompatible software stacks or browser sandboxes coexist without interference), \emph{model awareness} (different multi-agent roles route to separate trainable policies), and \emph{operational persistence} (clients can attach, detach, or fail without interrupting the central optimizer). Together, these properties map the limitations above onto explicit swarm topologies: model awareness addresses multi-model constraints through multiple trainable servers, runtime isolation addresses fragile and incompatible environments through detachable clients, operational persistence reduces iteration overhead and enables fault recovery, and server-side context tracking addresses redundant context without instrumenting the agent. The agent side can iterate rapidly while the training side remains stable. Making this separation efficient and trainable requires further mechanisms.
We organize the system contribution around three components. The first two constitute the core training substrate, while the third shows how the same substrate extends to automated experiment operation.

\begin{enumerate}
\item \textbf{Topology-configurable training} (Sections~\ref{sec:agentjet} and~\ref{sec:experiments}).
A single AgentJet training network can host many swarm servers (model optimizers) and many swarm clients (agent rollout centers) at once. The training paradigm is therefore expressed by the topology connecting servers and clients, rather than being fixed by a centralized trainer implementation. By reconfiguring this topology, AgentJet realizes each of the following more easily than centralized frameworks:
\begin{itemize}
\item Multiple swarm servers + a single swarm client: heterogeneous multi-model RL;
\item A single swarm server + multiple swarm clients: mixed-task training over isolated runtimes;
\item Swarm servers + dynamically hot-swappable swarm clients: REPL-style live code iteration;
\item Swarm servers + load balancing across multiple swarm clients: fault-tolerant client replacement.
\end{itemize}

\item \textbf{Efficient, black-box training over the serving API} (Sections~\ref{sec:pool_full} and~\ref{sec:blackbox}). AgentJet records LLM calls at OpenAI Chat/Responses-compatible endpoints with per-episode context tracking, organizes completed episodes through sample-pool-driven batching strategies, and applies \emph{timeline merging} to remove redundant multi-turn context. Timeline merging is configurable: more aggressive settings accelerate training but risk training-inference inconsistency, whereas the conservative default preserves the observed rollout behavior. On an AppWorld workload, this default reduces actor-update time by 6.25$\times$ on average.

\item \textbf{AutoResearch for long-horizon experiment operation} (Section~\ref{sec:auto_research}). Given a research topic as input, autonomous agents use the same swarm interface to plan, launch, monitor, debug, and synthesize multi-day RL experiment campaigns on large-scale GPU clusters, reducing the amount of manual orchestration required from RL researchers.
\end{enumerate}

Beyond these core mechanisms, Section~\ref{sec:experiments} evaluates AgentJet across the swarm topology variants: shared- and non-shared-parameter multi-agent training in Werewolves, pipeline-internalized academic translation, mixed-task RL over AppWorld and AIME with isolated runtimes, and multi-turn training with timeline merging and framework-agnostic clients. Section~\ref{sec:auto_research} then shows that the same detachable-client design can support long-horizon AutoResearch campaigns. Together, these studies examine how policy separation, runtime isolation, and topology reconfiguration affect generalist and specialist agent behaviors.

AgentJet is fully open-source and supports agents implemented with the OpenAI SDK, LangChain, AgentScope, raw HTTP clients, and any scaffold that can issue compatible LLM inference requests.

\section{Related Work}
\label{sec:related_work}

\subsection{Reinforcement Learning Frameworks for LLM Agents}

Reinforcement learning has become a central mechanism for improving reasoning, tool use, and agentic behavior in large language models, and recent surveys characterize agentic RL as a shift from single-step text generation to temporally extended interaction in which policies operate through tools, environments, memory, and delayed feedback~\citep{zhang2025agenticrlsurvey}. We organize the resulting landscape along two axes: general-purpose RLHF systems and agent-native RL systems.

\paragraph{General-Purpose RLHF Frameworks.}
OpenRLHF~\citep{hu2024openrlhf} pioneered a Ray~\citep{moritz2018ray} + vLLM~\citep{kwon2023vllm} distributed architecture that orchestrates actor, reward, reference, and critic models across GPU clusters with hybrid engine scheduling, while veRL (HybridFlow)~\citep{sheng2024hybridflow} introduced a hybrid single-controller/multi-controller programming model that decouples RL control flow from computation flow, achieving state-of-the-art throughput across PPO~\citep{schulman2017ppo}, GRPO~\citep{shao2024deepseekmath}, and other algorithms. These frameworks are essential optimizer-side infrastructure and have been instrumental in scaling reasoning RL, but they were designed primarily for single-turn or short-horizon interactions and typically assume that rollout generation, reward computation, and model updates are organized within a single tightly coupled training job.

\paragraph{Agent-Native RL Frameworks.}
A new wave of frameworks relaxes this assumption in different ways. Forge by MiniMax~\citep{minimax2026forge} introduces a middleware abstraction that treats agents as black boxes whose LLM requests are routed through a service gateway, and it integrates Context Management as an explicit agent action within the RL loop to mitigate context rot. AReaL~\citep{fu2025areal} pioneered fully asynchronous training--inference decoupling, achieving a 3--5$\times$ speedup in agent search scenarios, and its Proxy Worker middleware enables any agent to connect to RL training with a single API-endpoint change; ROLL Flash~\citep{lu2025rollflash} similarly decouples generation from training to improve utilization under asynchronous rollout. AgentRL~\citep{zhang2025agentrl} targets multi-turn, multi-task agentic RL with asynchronous generation, heterogeneous environment deployment, and task-level normalization, and SkyRL-Agent~\citep{cao2025skyrla} emphasizes efficient multi-turn rollout orchestration, tool integration, and backend interoperability. Agent Lightning~\citep{luo2025agentlightning} converts agent experience into state--action--reward transitions for low-intrusion integration with existing frameworks, OpenTinker~\citep{zhu2026opentinker} advances the separation-of-concerns paradigm with a Reinforcement-Learning-as-a-Service architecture and an Agent Protocol Coordinator for multi-agent training, and OpenClaw-RL~\citep{wang2026openclawrl} treats every agent interaction as a live training signal for continuous online policy optimization.
The slime framework~\citep{thudm2025slime} further emphasizes RL scaling through Megatron-LM training and SGLang-based rollout generation. 

AgentJet builds on this line of work but targets a different systems gap. Existing frameworks increasingly recognize the need to separate agent execution from optimization, yet the interface between arbitrary agent runtimes and persistent trainable model servers remains fragmented. AgentJet treats standard LLM serving requests as the common connection point: OpenAI Chat/Responses-compatible calls are routed with per-episode credentials, captured by server-side context trackers, and admitted into task-organized sample pools. This makes the serving path itself responsible for linking agent execution, reward attribution, trainable policies, and optimizer state, while leaving the client-side agent largely unchanged, yielding a many-to-many, fault-tolerant topology rather than a new optimization algorithm or middleware.

\subsection{Multi-Agent Systems and Multi-Agent Reinforcement Learning}

Multi-agent systems study how multiple agents coordinate, communicate, specialize, or compete. In the LLM setting, frameworks such as AutoGen~\citep{wu2023autogen}, AgentScope~\citep{gao2024agentscope}, MetaGPT~\citep{hong2024metagpt}, and CrewAI~\citep{crewai} provide abstractions for role assignment, message passing, tool use, and collaborative workflows. These systems make it easier to build multi-agent applications, but their primary concern is agent orchestration rather than reinforcement learning over multiple trainable policies.

Multi-agent reinforcement learning (MARL) studies a complementary problem: how multiple learning policies should be optimized in shared, cooperative, competitive, or mixed-motive environments. Classical MARL algorithms address non-stationarity, credit assignment, and centralized-training/decentralized-execution trade-offs, for example through multi-agent actor--critic methods~\citep{lowe2017multi} or value-factorization approaches such as QMIX~\citep{rashid2018qmix}. These methods establish important algorithmic foundations, but most are not designed for LLM agents whose actions are multi-turn language/tool trajectories and whose execution environments may be implemented by arbitrary agent frameworks.

AgentJet connects these two lines. It neither replaces LLM multi-agent frameworks nor proposes a new MARL algorithm; instead, it provides a system topology in which multiple trainable LLM policies can be hosted by independent swarm servers while one or more clients run the shared game, workflow, reward, or evaluator. This is the gap targeted by our shared- and non-shared-parameter Werewolves studies: AgentJet allows the same client-side multi-agent environment to switch between a shared policy and multiple independently optimized policies without rewriting the environment or reward protocol. The academic translation study further shows that a multi-agent proposal--review--modify workflow can act as a training scaffold for a single deployable policy.

\subsection{Multi-Task and Heterogeneous Agent Training}

Multi-task learning studies how a model can learn across tasks by sharing representations or policy structure~\citep{caruana1997multitask}. In reinforcement learning, systems such as Distral~\citep{teh2017distral} and IMPALA~\citep{espeholt2018impala} show how multi-task or distributed actor--learner training can improve transfer and utilization, while also revealing challenges such as negative transfer, reward-scale imbalance, and task interference. For LLM agents, these challenges become more operational: tasks may require incompatible packages, browser stacks, external APIs, MCP services~\citep{mcp2024}, verifiers, or simulators, as illustrated by the diversity of interactive runtimes in benchmarks such as SWE-bench~\citep{jimenez2024swebench}, WebArena~\citep{zhou2024webarena}, Mind2Web~\citep{deng2023mind2web}, and AppWorld~\citep{trivedi2024appworld}.

Recent agentic RL frameworks have begun to address this problem: AgentRL explicitly studies multi-turn, multi-task agent training and introduces environment APIs and normalization strategies for heterogeneous tasks~\citep{zhang2025agentrl}, and SkyRL-Agent demonstrates backend interoperability and tool-enhanced training recipes across long-horizon agent settings~\citep{cao2025skyrla}. However, many systems still organize task execution around a centralized trainer, controller, or framework-specific environment abstraction, which can make it difficult to combine tasks whose dependencies are mutually incompatible or whose evaluation logic must evolve independently from the optimizer.

AgentJet frames heterogeneous task training as a client--server composition problem. Each task runtime can remain inside its own swarm client, container, virtual environment, or machine, while completed episodes flow into persistent trainable servers through the same serving-layer data path; sample-pool-driven batching then organizes episodes by task identity or synchronization condition before policy updates. This design motivates our mixed-task RL study over AppWorld and AIME: the goal is not to claim that joint training always dominates specialized training, but to show that isolated runtimes can feed a shared optimization run and expose the trade-off between deployable generalists and task-specific specialists.

\subsection{Automated Scientific Research}

The vision of AI-conducted research has advanced rapidly. AI Scientist from Sakana AI~\citep{lu2024aiscientist} demonstrated end-to-end automated research, from idea generation through code implementation, experimentation, paper writing, and peer review. AI Co-Scientist from Google DeepMind~\citep{lu2025nature} further explored AI-driven scientific hypothesis generation and experimental design. AI-Researcher~\citep{tang2024airesearcher} from HKU received a NeurIPS 2025 Spotlight, and AgentRxiv~\citep{schmidgall2025agentrxiv} employs multi-agent collaboration to iteratively improve results, increasing MATH-500~\citep{hendrycks2021math,lightman2024verify} accuracy from 70.2\% to 78.2\% through automated research cycles. OpenAI has also identified fully automated AI researchers as a core long-term objective~\citep{openai2026researcher}. These systems mainly study the intelligence, coordination, and evaluation of research agents and predominantly focus on producing research papers or optimizing benchmark scores on tasks that complete in minutes.

AgentJet studies a complementary infrastructure problem: how can research agents operate long-running RL campaigns without becoming part of the optimizer itself? The AutoResearch pipelines of AgentJet target the less-explored problem of orchestrating \emph{long-horizon experimental campaigns} in which individual training runs last hours to days, multiple experiments must be scheduled across GPU clusters, and adaptive multi-stage experimental design is required to reach research conclusions. Throughout multi-day training, experiment controllers may revise hypotheses, launch follow-up runs, recover failed workers, or inspect logs while model weights, optimizer state, and sample pools must remain stable. AgentJet keeps this separation explicit: AutoResearch agents run as clients that control experiments through the same serving-layer training interface used by ordinary rollout clients, while persistent swarm servers maintain training state. This identifies a research gap not fully addressed by prior automated research systems, namely connecting autonomous experiment operation to robust, recoverable, and persistent RL training infrastructure.

\section{Preliminaries}
\label{sec:preliminaries}

This report concerns reinforcement learning (RL) of LLM-based agents. We work with verifiable or executable rewards, instantiated through GRPO-style policy-gradient updates~\citep{shao2024deepseekmath}.

Agentic RL instead optimizes over the full execution of an agent workflow. We describe it through four nested concepts:
\begin{itemize}
  \item An \textbf{episode} is the complete execution of an agentic task, e.g. solving a math problem.
  \item A \textbf{turn} is a single API-level LLM request/response event within an episode, counted per trainable policy rather than per agent. Over one episode, an agent may issue many turns, invoke tools, and mutate external state before any reward is observed. In a multi-agent episode where several agents share the same trainable policy, each agent's individual request is a turn of that shared policy, so the policy may accumulate many turns across agents even when no single agent issues more than one.
  \item A \textbf{message} is an array of tokens that begins with a beginning-of-sequence (BoS) token and ends with an end-of-sequence (EoS) token. In agentic RL, we tag each message with its producer (e.g. \texttt{user}, \texttt{llm}, or \texttt{env}), so that masks and rewards can be assigned flexibly on a per-token basis.
  \item A \textbf{sample} is a token sequence with masks and other necessary metadata that is used for policy optimization; a typical sample's token sequence begins with input prompt tokens (\texttt{user} message), followed by alternating LLM-generated tokens (\texttt{llm} message) and environment feedback tokens (\texttt{env} message). Each episode may produce multiple samples when multiple turns take place in an episode.
\end{itemize}

The episode-turn-message-sample distinction matters because agents query the same trainable policy repeatedly within an episode, appending earlier turns as context. A trainer that treats each episode as a flat sequence will either misattribute rewards to the wrong turns or waste computation retraining on duplicated prefix tokens. A precise mapping from episodes to turns to messages to samples is therefore essential for efficient agentic RL. The next section turns this data model into system machinery: swarm servers use per-episode routing credentials to reconstruct turns from serving requests, organize the resulting samples into task-aware pools, and merge redundant timelines before optimization.

\section{AgentJet: Design and Architecture}
\label{sec:agentjet}

AgentJet is a distributed framework designed specifically for agentic reinforcement learning (RL).
It operationalizes the episode-turn-message-sample data model from Section~\ref{sec:preliminaries} through a serving-layer protocol: agent execution remains on client nodes, whereas model state, context reconstruction, sample pooling, and optimization remain on server nodes.
This infrastructure coordinates heterogeneous devices, with or without GPUs, to train one LLM model or multiple LLM models simultaneously, thereby improving task performance and model capabilities.
AgentJet adopts a swarm architecture in which multiple nodes form a training network for a wide range of agentic training scenarios.

\subsection{Swarm Architecture}
\label{sec:system_architecture}

\begin{figure}[htbp]
\centerline{\includegraphics[width=\linewidth]{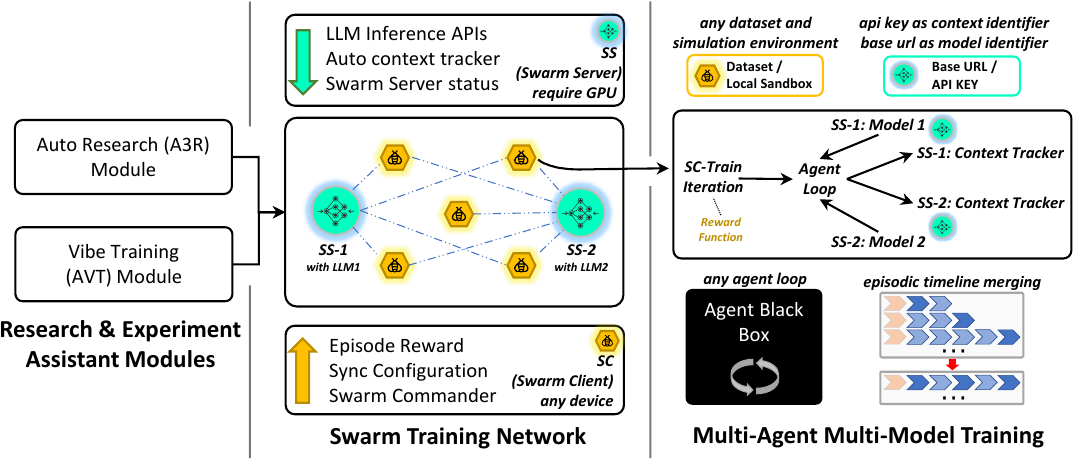}}
\caption{AgentJet swarm training architecture. The swarm servers (optimizer nodes) host model weights on GPU clusters, while the swarm clients (sampling nodes) execute agent workflows and collect reinforcement learning training trajectories. The swarm network is formed dynamically as clients join and leave, enabling fault-tolerant operation, hot-swap debugging, and multi-model training.}
\label{fig:system_arch}
\end{figure}

The majority of existing agentic RL frameworks rest on two implicit coupling assumptions. First, all agents within a given task are required to share a single trainable LLM, as prevailing training backends are designed to optimize a single model at a time. Second, rollout workers are tightly coupled to the training process, executing on the same machine and within the same runtime that hosts gradient computation. These assumptions conflate two workloads whose reliability profiles differ substantially: gradient computation is GPU-bound and benefits from tight co-location, whereas agent rollout is dominated by interactions with external tools, MCP~\citep{mcp2024} services, and remote resource access, whose routine failures (e.g., exhausted quotas, rate limits) frequently interrupt training and incur the loss of unsaved progress.

This coupling further precludes a number of research directions of growing importance, including heterogeneous multi-agent RL with non-shared parameters, multi-domain joint training, and on-policy training of agents that depend on heavyweight runtimes. AgentJet addresses both couplings by decoupling the training and rollout planes into a swarm of cooperating nodes.

As illustrated in Figure~\ref{fig:system_arch}, AgentJet employs a swarm-based training architecture orchestrated through two types of nodes, which can be deployed across one or multiple hardware devices, and which may join or leave the swarm at any time:

\begin{itemize}
    \item \textbf{Swarm Server Nodes (Optimizer Nodes).} Each server node is a training engine running on a GPU server or cluster. Server nodes execute LLM policy gradient updates, host vLLM~\citep{kwon2023vllm}/SGLang~\citep{zheng2024sglang} inference APIs with automatic context tracking, and manage episode lifecycles. Multiple server nodes can operate concurrently, each serving and training a distinct model (e.g., Qwen3-32B and Qwen3-14B), enabling non-shared-parameter multi-agent training. The current server node implementation is verl-based~\citep{sheng2024hybridflow} for best compatibility with most RL algorithms.

    \item \textbf{Swarm Client Nodes (Sampling Nodes).} Client nodes are lightweight, CPU-only processes that can run on any device, including workstations, laptops, or the same machines that host servers. Each client executes arbitrary agent workflows, reads datasets, runs agent loops, and computes task-specific rewards. As clients interact with the inference API of a server, the server transparently captures all context for training. Moreover, clients can be authorized to manipulate server nodes to update training configurations, terminate or restart training, and fetch training progress. Client nodes can also restart and rejoin the training network safely without affecting the overall training process.
\end{itemize}

An AgentJet training network (swarm) is composed of server and client nodes that are interconnected yet independently operated.
On the one hand, the number of server nodes is determined by the number of models being trained, since each server independently hosts and trains a distinct model.
On the other hand, the number of client nodes is highly flexible; researchers can dynamically add nodes to increase redundancy, train on multiple datasets simultaneously, run unscheduled evaluations, debug reward functions mid-training, and so on.
Individual clients can be freely terminated, modified, and restarted while the rest of the swarm continues training uninterrupted, and different clients may run in entirely isolated runtime environments, such as separate Docker containers, virtual machines, or even distinct operating systems.
These two degrees of freedom---the number of trainable servers and the number of isolated clients---make topology a system-level design variable: changing the topology changes the training regime while preserving the same serving-optimization interface.
Section~\ref{sec:experiments} evaluates this topology-configurable design across several training settings.

\subsection{Swarm Reinforcement Learning Paradigm}
\label{sec:interaction}

\begin{figure}[htbp]
\centerline{\includegraphics[width=0.9\linewidth]{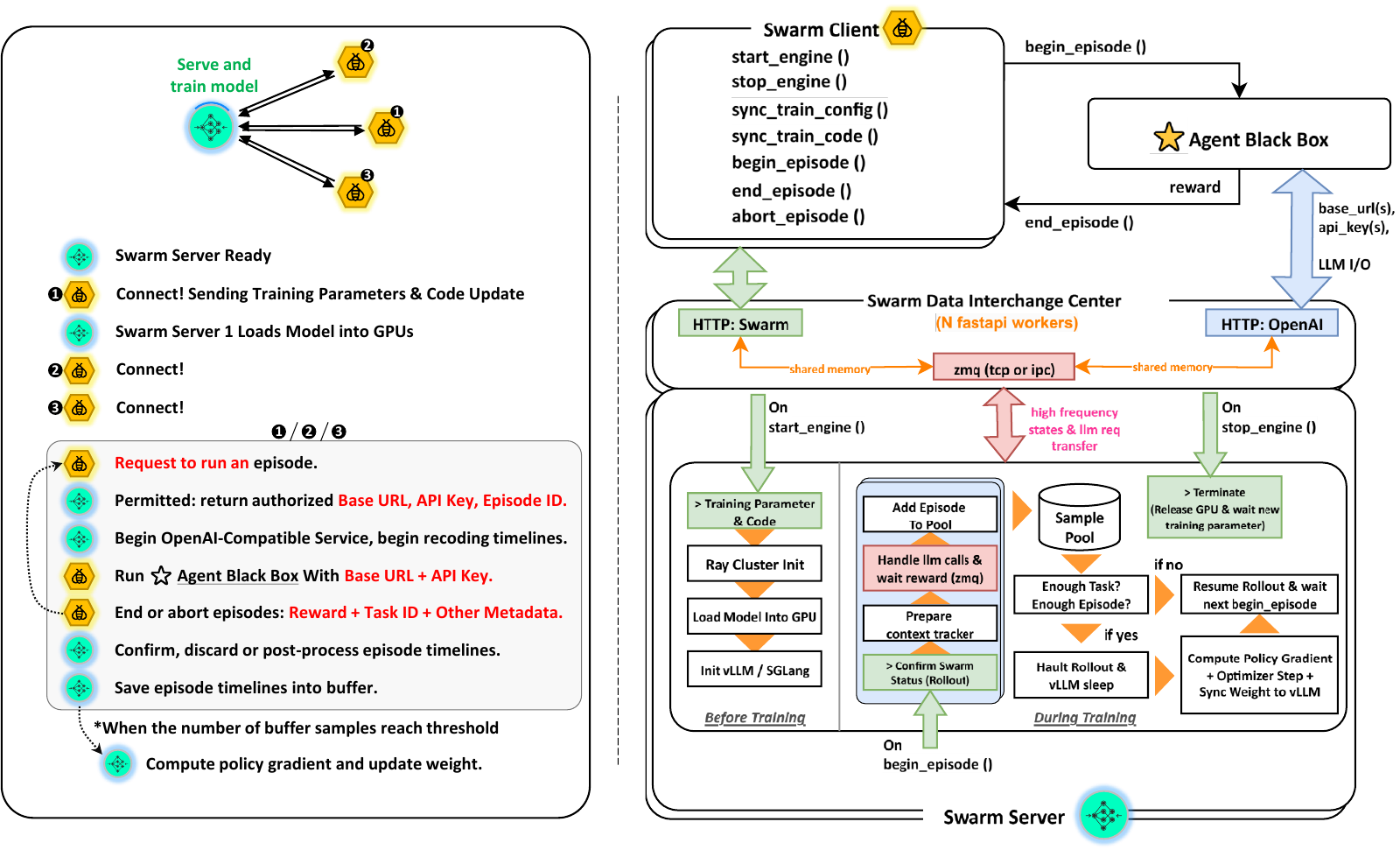}}
\caption{Swarm reinforcement learning paradigm between server and client nodes, showing network establishment, episode lifecycle, and asynchronous weight update trigger.}
\label{fig:communication}
\end{figure}

The reinforcement learning procedure in AgentJet is conceptually similar to that of other RL frameworks; what differs is how responsibilities are distributed across the swarm.
By decoupling trajectory collection from model optimization, the swarm architecture makes the RL process more flexible and scalable, and accommodates more sophisticated experimental setups.
The basic procedure unfolds in the following phases:

\paragraph{(1) Setting up swarm servers.}
The swarm servers carry models and model optimizers in the RL training swarm network.
When initialized, each swarm server starts from scratch (without loading any training parameters or LLM models) and only claims an address to register itself in the swarm network, waiting to be woken by a swarm client that supplies the training configuration and the booting instruction.

\paragraph{(2) Connecting swarm clients.}
Swarm clients are agent runners and RL trajectory generators that collect agent trajectories by executing agents using the LLM designated for training.
A swarm can include an arbitrary number of swarm clients, ranging from a single client to many, running on any set of devices and operating systems.
In either case, one designated client acts as the controller, which activates and deactivates swarm servers and synchronizes training parameters.
Once the training parameters have been delivered to the swarm servers, the swarm servers enter the \texttt{Booting} phase, and finally, the \texttt{Rolling} phase when all model weights are loaded into memory.

\paragraph{(3) Collecting RL Samples.}
Reinforcement learning commences in this phase.
Swarm servers expose an OpenAI-compatible inference API for swarm clients to execute agent episodes (i.e., black-box agent loops or dynamic workflows).
This process transparently captures all context and trajectories for subsequent policy training.
To initiate an episode, a swarm client first registers the episode at the participating swarm servers.
Upon approval, each swarm server returns a dedicated \texttt{BaseUrl}/\texttt{ApiKey} pair,
which is encoded with model and episode metadata to avoid collision with other episodes running in parallel.
Concurrently, each swarm server also creates an internal \textbf{context tracker} associated with the episode to record all agent LLM requests routed through it.
Upon episode termination, the swarm client has two choices: 
either submit a \textbf{reward signal} and allow the episode samples to enter the \textbf{sample pools} in the swarm server,
or submit an \textbf{abort signal} to instruct the swarm server to discard this episode.

\paragraph{(4) Optimizing the model.}
When the sample pool is full or other predefined conditions are met (refer to Sec.~\ref{sec:pool_full}), 
the swarm server transitions into the \texttt{Weight\_Syncing} phase to optimize the LLM. 
First, 
the timeline merging module of AgentJet merges trajectories from the sample pool to reduce redundancy. 
Second, 
policy gradient and other loss terms are computed to perform optimizer steps in accordance with the chosen RL algorithm, 
updating either the full model weights or the LoRA~\citep{hu2021lora} matrix weights. 
Finally, 
the swarm server returns to the \texttt{Rolling} phase to resume trajectory collection as described in step (3).

\paragraph{(5) Evaluation.}
One advantage of the swarm architecture is that evaluation tasks can be scheduled at any time and on any node.
Training and evaluation can proceed simultaneously; the key distinction is that evaluation episodes must be aborted upon completion to prevent their trajectories from entering the training sample pool.

\subsection{Dynamic Batching}
\label{sec:pool_full}

Most traditional RL frameworks adopt pre-scheduled batching behavior, assuming that the episode batch size, 
group volume, and task composition remain fixed throughout the entire training process. 
However, this practice hinders the flexible utilization of advanced sampling control methods, 
such as dynamic sampling in DAPO~\citep{yu2025dapo}, 
where post-hoc filtering is applied to discard samples with zero advantage estimates,
a non-trivial procedure that is at odds with pre-scheduled batching.
AgentJet's swarm-based architecture instead enables flexible options to dynamically control sample batching, 
making it notably straightforward to blend heterogeneous samples and improve sampling efficiency.

\begin{figure}[htbp]
\centerline{\includegraphics[width=0.95\linewidth]{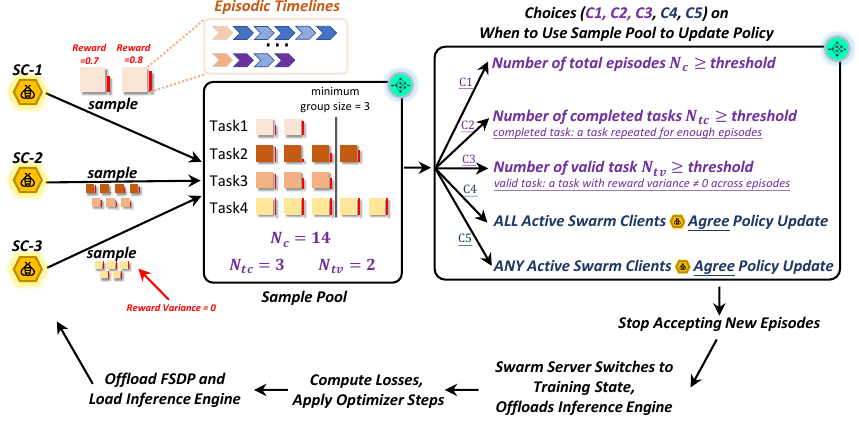}}
\caption{Swarm-coordinated reinforcement learning in AgentJet.
Swarm clients run agent episodes against OpenAI-compatible inference endpoints exposed by swarm servers; completed episodes with rewards are accumulated in a task-organized sample pool.
Furthermore, the sample pool collecting strategy can be adjusted according to specific experimental requirements.
}
\label{fig:swarm_coordinated_rl}
\end{figure}

As the episodes complete, the server accumulates them in a sample pool organized by task identity. 
Let $\mathcal{B} = \{(t_i, \mathcal{E}_i)\}$ denote the pool, where $t_i$ is a unique task identifier and $\mathcal{E}_i = \{e_i^1, \dots, e_i^{|\mathcal{E}_i|}\}$ is the set of completed episodes for task $t_i$, each with reward $r_i^j$. 
Let $B$ denote the batch size and $N$ the number of rollout repeats per task.
AgentJet provides five collection strategies, organized into two groups: \emph{server side strategies} (C1--C3) and \emph{server-client agreement strategies} (C4--C5).
The former (C1--C3) adopts a pure producer-consumer model, in which the server unilaterally determines when the sample pool $\mathcal{B}$ is ready to trigger the next policy update.
The latter (C4--C5) incorporates the status of client nodes into the scheduling decision; specifically, the server waits for acknowledgment from one or all client nodes before proceeding with the next policy update.

\begin{itemize}
    \item \textbf{C1. Episode-count strategy.} Let $N_c = \sum_{i} |\mathcal{E}_i|$ be the total number of completed episodes. Training is triggered when $N_c \geq B \times N$.

    \item \textbf{C2. Task-count strategy.} A task $t_i$ is considered \emph{completed} when $|\mathcal{E}_i| \geq N$ (i.e., it has accumulated a sufficient number of episode repetitions as a group). Let $N_{tc} = |\{t_i : |\mathcal{E}_i| \geq N\}|$ denote the number of completed tasks. Training is triggered when $N_{tc} \geq B$.

    \item \textbf{C3. Non-dummy task-count strategy.} A \emph{completed} task is further classified as \emph{valid} (non-dummy) only if its episodes exhibit non-zero reward variance (i.e., $\exists\, j \neq k$ s.t.\ $r_i^j \neq r_i^k$). Let $N_{tv}$ denote the number of valid tasks; training is triggered when $N_{tv} \geq B$. This dynamic sampling strategy ensures that all collected samples provide reliable policy gradient signals, at the cost of increased sampling time.

    \item \textbf{C4. All-clients-agree strategy.} Training is triggered when all active swarm client nodes have signaled agreement. This strategy is appropriate when the proportion of samples contributed by different client nodes should be controlled, or when fine-grained coordination among client nodes is required in heterogeneous multi-agent cooperation scenarios.

    \item \textbf{C5. Any-client-agrees strategy.} Training is triggered as soon as any active swarm client node signals agreement.
\end{itemize}

Once a policy update is triggered, a swarm server node first transitions into the \texttt{Rolling\_Post} state, in which new episode claims are rejected while in-flight episodes are allowed to drain. It then advances to the \texttt{Weight\_Syncing} state, where the policy gradient step is performed and the updated weights are broadcast to the inference engine. As soon as all the optimizer steps are performed, the server node switches back to \texttt{Rolling} state.


\begin{figure}[htbp]
\centering
\begin{minipage}[t]{0.37\linewidth}
    \centering
    \includegraphics[width=\linewidth]{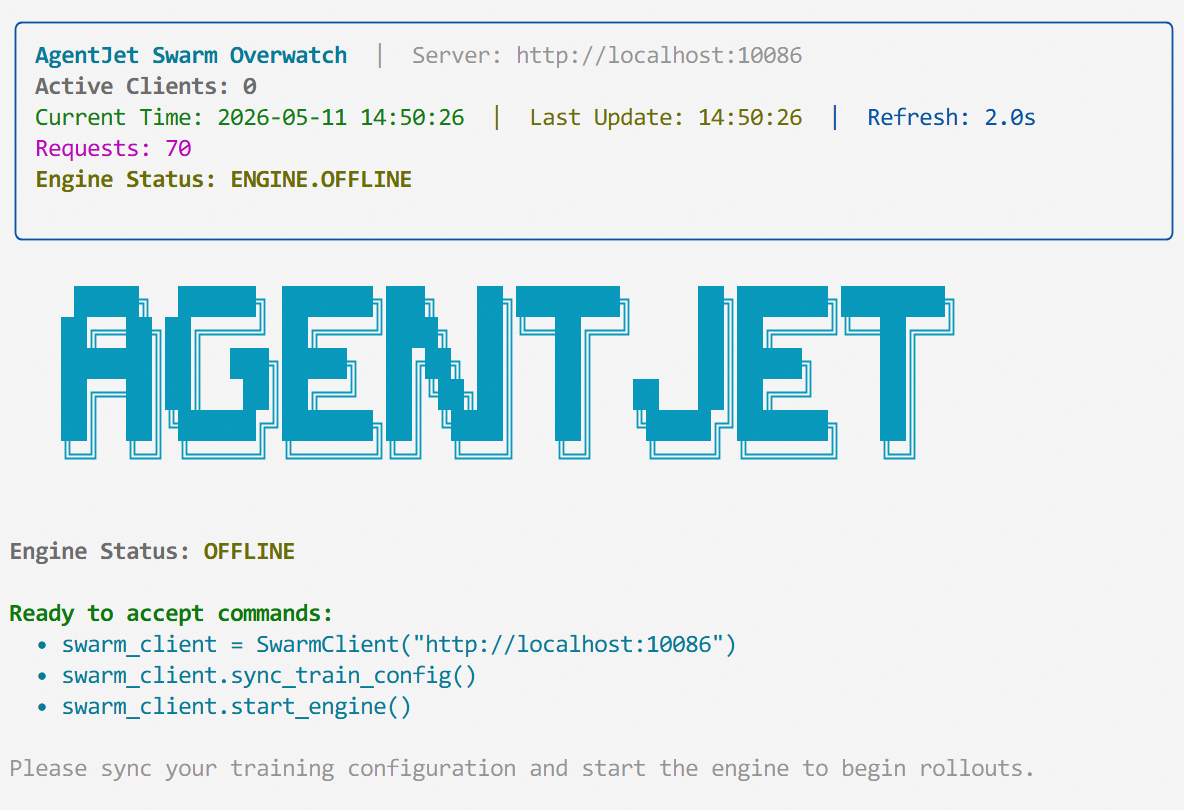}
\end{minipage}
\hfill
\begin{minipage}[t]{0.61\linewidth}
    \centering
    \includegraphics[width=\linewidth]{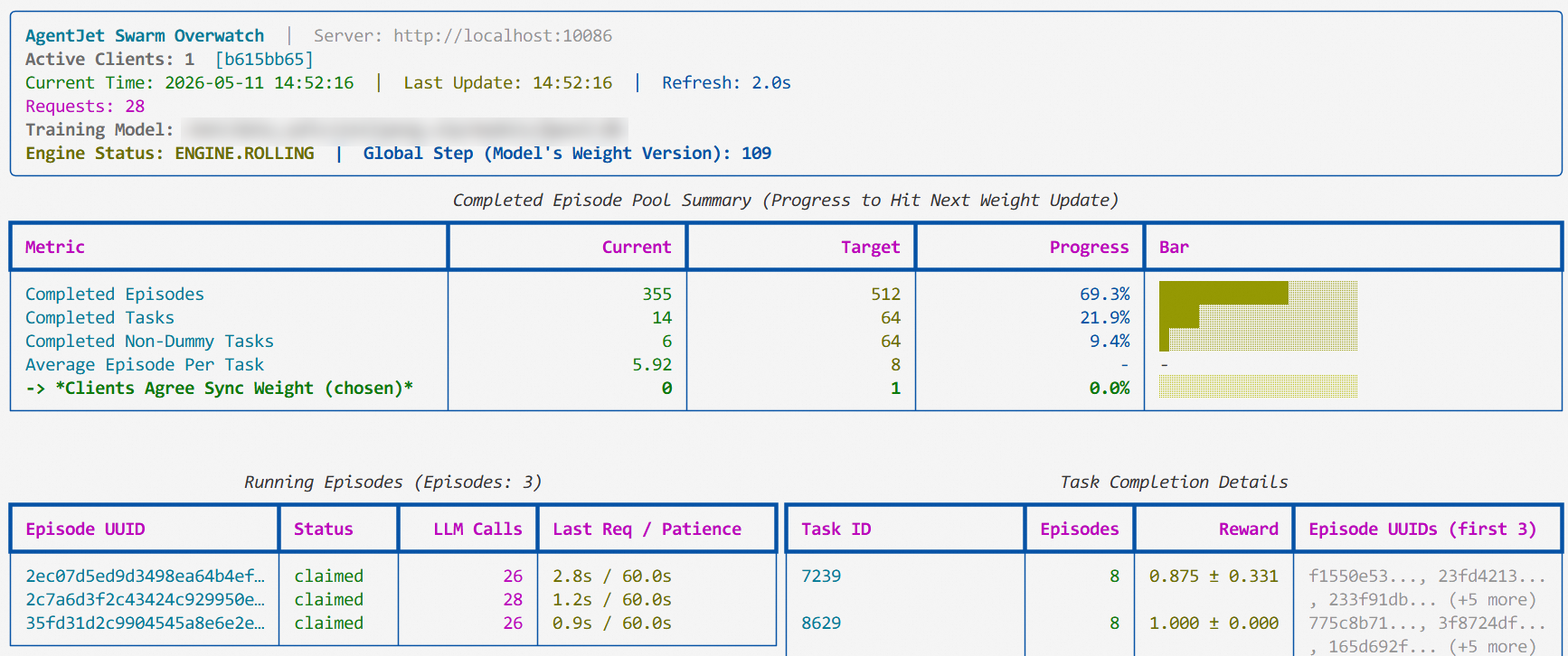}
\end{minipage}
\caption{Swarm server monitoring with live sample-collection progress driving the C1--C5 trigger criteria of Figure~\ref{fig:swarm_coordinated_rl}.}
\label{fig:overwatch}
\end{figure}

\subsection{Context Tracking and Framework-Agnostic Agent Support}
\label{sec:blackbox}

Many modern agents adopt sophisticated internal designs that manage the LLM context through mechanisms such as memory, skills, goal loops, and automatic compaction. To improve LLM performance via reinforcement learning within such agents, the training framework must be framework-agnostic and provide support for \emph{black-box agents}, that is, agent implementations that require \textbf{zero modification} and no framework-level instrumentation. AgentJet achieves this through an episode-level context tracking and reconstruction mechanism that operates independently within each swarm server.

\subsubsection{Framework-Agnostic Integration}
To maximize portability across the heterogeneous ecosystem of agent frameworks, each swarm server simultaneously exposes inference endpoints conforming to three widely adopted API protocols: the classic OpenAI Chat Completions API, the OpenAI Responses API, and the Anthropic Messages API.
By presenting a protocol-compatible surrogate at the network boundary, AgentJet enables any agent built atop these standards to be trained without source modification: the practitioner simply redirects the agent's configured LLM base URL and API key to the endpoints served by AgentJet. The API key is episode-specific and encoded with episode metadata; see Sec.~\ref{sec:interaction}.
Irrespective of the protocol employed, all inference requests are uniformly intercepted, tokenized, logged, and forwarded to the underlying vLLM/SGLang backend, rendering the choice of API surface orthogonal to context tracking.
Consequently, AgentJet supports any agent framework that communicates over HTTP, including LangChain~\citep{langchain}, AgentScope~\citep{gao2024agentscope}, MetaGPT~\citep{hong2024metagpt}, AutoGen~\citep{wu2023autogen}, CrewAI~\citep{crewai}, the OpenAI and Anthropic SDKs, and custom HTTP implementations.

\subsubsection{Context Tracker and Timeline Merging}
\label{sec:timeline_merging}

\paragraph{Context Tracker.} 
The context tracker is responsible for recording language model activities associated with an episode, 
and eventually producing the RL samples with masks, rewards, etc.
Each intercepted LLM call, corresponding to one turn in the terminology of Section~\ref{sec:preliminaries}, produces an independent \emph{timeline} in the context tracker.
Each timeline is an array of message blocks, where \textbf{each block} contains:
    an author label (\texttt{llm}, \texttt{env}, \texttt{user}),
    a piece of text,
    a token ID sequence,
    a per-token log probability sequence,
    and a binary loss mask sequence.
Formally, a timeline $T_i$ for the $i$-th LLM call is an ordered array of $K_i$ message blocks:
\begin{equation}
    T_i = \big(B_i^{[1]}, B_i^{[2]}, \dots, B_i^{[K_i]}\big), \qquad B_i^{[k]} = \big(\mathbf{m}_i^{[k]},\; \mathbf{x}_i^{[k]},\; a_i^{[k]},\; \boldsymbol{\ell}_i^{[k]},\; \boldsymbol{\mu}_i^{[k]}\big),
\end{equation}
where $\mathbf{m}_i^{[k]}$ is the text message, $\mathbf{x}_i^{[k]}$ the token ID sequence, $a_i^{[k]} \in \{\texttt{llm}, \texttt{env}, \texttt{user}\}$ the author label, $\boldsymbol{\ell}_i^{[k]}$ the per-token log probabilities, and $\boldsymbol{\mu}_i^{[k]} \in \{0, 1\}^{|\mathbf{x}_i^{[k]}|}$ the loss mask (set to 1 only for LLM-generated tokens).

\paragraph{Timeline Merging.}
Upon episode termination, the context tracker holds an ordered list $\mathcal{T} = (T_1, \dots, T_n)$ of timelines, one per intercepted LLM call. 
Because later turns reuse earlier turns as prompt, these timelines share long common prefixes: naively processing each as an independent training sample would re-process the same growing prefix on every call, incurring high computational overhead in the actor update. 
Timeline merging absorbs each timeline that matches a prefix of a longer one into that longer one, preserving which tokens were LLM-generated. 
Concretely, merging proceeds in reverse order (Algorithm~\ref{alg:timeline-merge}): $T_n$ absorbs each of $T_{n-1}, \dots, T_1$ in turn, then $T_{n-1}$ (if not yet absorbed) absorbs the remaining earlier timelines, and so on. 
The surviving timelines form $\mathcal{T}'$.

\begin{algorithm}[H]
\caption{Timeline Merging}
\label{alg:timeline-merge}
\textbf{Pairwise mergeability.} A longer timeline $S$ and shorter timeline $T$ are \emph{mergeable} when $|S| \ge |T|$ and $\text{Match}(S^{[k]}, T^{[k]}) = \text{True}$ for all $k \in [1, |T|]$, with $\text{Match}$ being token- or text-level equality per the chosen strategy (see Relaxed Matching Strategies below).

\begin{algorithmic}[1]
\REQUIRE Ordered timelines $\mathcal{T} = (T_1, T_2, \dots, T_n)$ in collection-time order; matching predicate $\text{Match}(\cdot, \cdot)$.
\ENSURE Merged timeline set $\mathcal{T}'$ with absorbed timelines removed.
\STATE $\mathcal{A} \gets \emptyset$ \hfill \COMMENT{indices of absorbed timelines}
\FOR{$i = n$ \textbf{downto} $2$}
    \IF{$i \in \mathcal{A}$}
        \STATE \textbf{continue}
    \ENDIF
    \FOR{$j = i-1$ \textbf{downto} $1$}
        \IF{$j \in \mathcal{A}$}
            \STATE \textbf{continue}
        \ENDIF
        \IF{$|T_i| \ge |T_j|$ \textbf{and} $\text{Match}\big(T_i^{[k]}, T_j^{[k]}\big), \forall k \in [1, |T_j|]$}
            \FOR{$k = 1$ \textbf{to} $|T_j|$}
                \IF{$T_j^{[k]}.\text{author} = \mathtt{llm}$ \textbf{and} $T_i^{[k]}.\text{author} \neq \mathtt{llm}$}
                    \STATE $T_i^{[k]}.\text{author} \gets \mathtt{llm}$
                    \STATE $T_i^{[k]}.\text{tokens} \gets T_j^{[k]}.\text{tokens}$
                    \STATE $T_i^{[k]}.\text{logprob} \gets T_j^{[k]}.\text{logprob}$
                \ENDIF
            \ENDFOR
            \STATE $\mathcal{A} \gets \mathcal{A} \cup \{j\}$ \hfill \COMMENT{$T_j$ absorbed into $T_i$}
        \ENDIF
    \ENDFOR
\ENDFOR
\RETURN $\mathcal{T}' = (T_i)_{i \notin \mathcal{A}}$ in original order
\end{algorithmic}
\end{algorithm}

\paragraph{Relaxed Matching Strategies.}
To handle real-world tokenization drift (text encoded then re-encoded produces different token sequences), two configurable matching strategy options are provided:
\begin{enumerate}
    \item \textbf{token}: exact token-id match at every aligned position; suited for strict training/inference consistency.
    \item \textbf{text} (default): exact text match at every aligned position; tolerates tokenizer drift and yields more aggressive merging.
\end{enumerate}

\begin{figure}[H]
\centering
\includegraphics[width=0.7\linewidth]{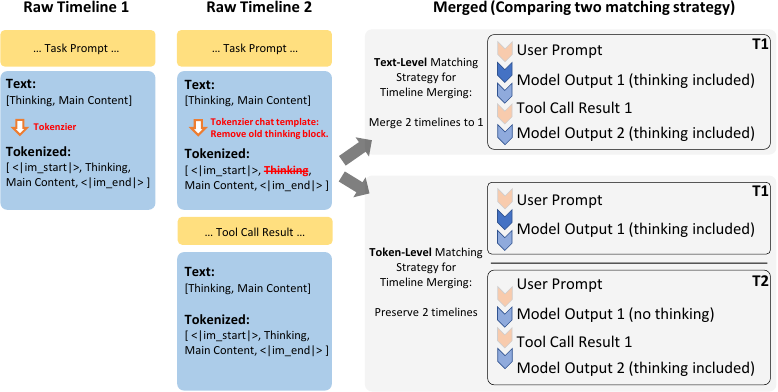}
\caption{Text-level vs.\ token-level timeline matching on Qwen3. The earlier assistant message has its thinking block stripped once a later turn is appended, so the two timelines agree on text but diverge in token ids: text-level merges them, token-level keeps them separate.}
\label{fig:text_vs_token_merge}
\end{figure}

For the majority of timeline pairs, the two strategies produce identical results, but they diverge precisely at the boundary where researchers must trade training efficiency against train-inference (TI) consistency.
Figure~\ref{fig:text_vs_token_merge} illustrates one such case: under the Qwen3 chat template, once a follow-up assistant turn is appended, the tokenizer strips the thinking block from earlier assistant messages, causing token array differences even if the text is the same.
Under \emph{text-level} matching, timelines merge as one, maximizing the training speed at the cost of TI consistency.
Under \emph{token-level} matching, timelines remain separated.
Token-level matching is preferred when the workload is sensitive to such tokenizer-induced drift and exact alignment with the inference-time token stream is required.

\subsection{REPL-Style Autonomous Research Capability}
\label{sec:repl}

AgentJet supports a \emph{Read--Eval--Print--Loop} (REPL) style research workflow directly against a live training run.

In conventional agentic RL frameworks,
the fast-changing components (blackbox agent loop, prompts, reward functions, and evaluation scripts) are usually co-located with the optimizer inside coupled single training processes.
Any edit to these components forces a full reboot:
reloading the model, reinitializing the inference engine, and discarding in-flight progress,
so that a single experimental tweak costs several to tens of minutes and evaluation must be pre-scheduled rather than probed on demand.

AgentJet eliminates this reboot cost by partitioning state along its reconstruction cost.
Swarm servers retain everything that is expensive to reconstruct (model weights, optimizer state, the inference engine, per-episode context trackers, and the accumulated sample pool), whereas swarm clients carry only the cheap, frequently revised logic---the black-box agent loop, prompts, reward functions, and evaluation scripts.
Because a client holds no irreplaceable state, it can be edited, terminated, detached, and reattached while the server keeps training uninterrupted.
Restarting a client costs only the time to reload its own code and environment, typically a few seconds, rather than the minutes required to reinitialize the optimizer.

This state partitioning allows the training topology to be reconfigured dynamically (Fig.~\ref{fig:swarm_topology}): 
when a black-box agent requires a quick fix, killing the old swarm client and attaching a patched one is sufficient.
The entire procedure takes only seconds, and any trajectory data abandoned by the discarded client is autonomously cleared or recycled without manual intervention.

Beyond dynamically modifying agents, dynamic reconfiguration of the training topology serves a number of other purposes; for example, it enables
(1)~adjusting the reward weighting in real time mid-training, e.g., adding or removing reward components, to interactively observe the model's response to reward changes;
(2)~reweighting the proportions of different tasks in real time mid-training, so that the overall training direction can be quickly corrected when multiple tasks are trained jointly;
(3)~attaching, without interrupting training, an additional swarm client backed by an entirely unrelated environment, to probe how generalization changes across benchmarks drawn from different domains;
and 
(4)~distributing the black-box agent across distinct devices via redundant nodes to improve system fault tolerance, among others.

In addition to the client-side and topology reconfigurations above, 
AgentJet supports live updates to the swarm server code as well.
A swarm client can issue commands that update the swarm server code running in the cloud GPU cluster on the fly,
enabling convenient agent-assisted RL algorithm development from anywhere.
More specifically, as shown in Fig.~\ref{fig:communication}, a swarm client can not only push new training parameters to the swarm server,
but also revise the code underlying any training algorithm and deploy the changes remotely;
through dedicated control commands, it can then instruct training to restart from scratch or resume from the most recent checkpoint.

\subsection{Live Debugging and Interactive Evaluation for Autonomous Research}
\label{sec:live_debugging}

Modern agent harness systems heavily depend on trial-and-error to rapidly convert ideas
into practical solutions and iteratively develop agent harnesses and rewards.

The same attachment mechanism lets a researcher---or, equivalently, an autonomous research agent---dynamically attach swarm clients for debugging and evaluation against a live run.
A fresh client can attach to an already-running server at any time to step through a failing case, inspect an intermediate trajectory, or probe whether the policy is forgetting an earlier capability; none of this restarts the optimizer.
Evaluation episodes are flagged for abort on completion,
so their trajectories never enter the training sample pool (Sec.~\ref{sec:interaction})
and interactive probes coexist with ongoing training.

Per-episode context tracking (Section~\ref{sec:blackbox}) is what allows client replacement to proceed without manual state recovery:
every inference request is intercepted and recorded server-side under an episode-specific credential (\texttt{ApiKey}),
so a restarted or freshly attached client immediately produces well-formed, capturable episodes with no client-side state to reconstruct.
This REPL substrate is the direct prerequisite for the autonomous-research capabilities of Section~\ref{sec:auto_research}:
an AI coding agent can generate or modify swarm clients, implement rewards, launch rollouts, inspect logs, and reattach in seconds,
iterating on failures without ever touching the server-side optimizer.

\section{Multi-Agent, Multi-Model and Multi-Task Training}
\label{sec:experiments}
\label{sec:architectural_advantages}

This section evaluates topology-configurable training in AgentJet. By varying the swarm training topology, the same server--client separation supports shared-parameter multi-agent training, non-shared multi-model training, mixed-task training, and efficient multi-turn training with timeline merging.
The swarm architecture separates GPU-intensive model training from agent harness logic and reward computation.

\begin{figure}[htbp]
\centerline{\includegraphics[width=\linewidth]{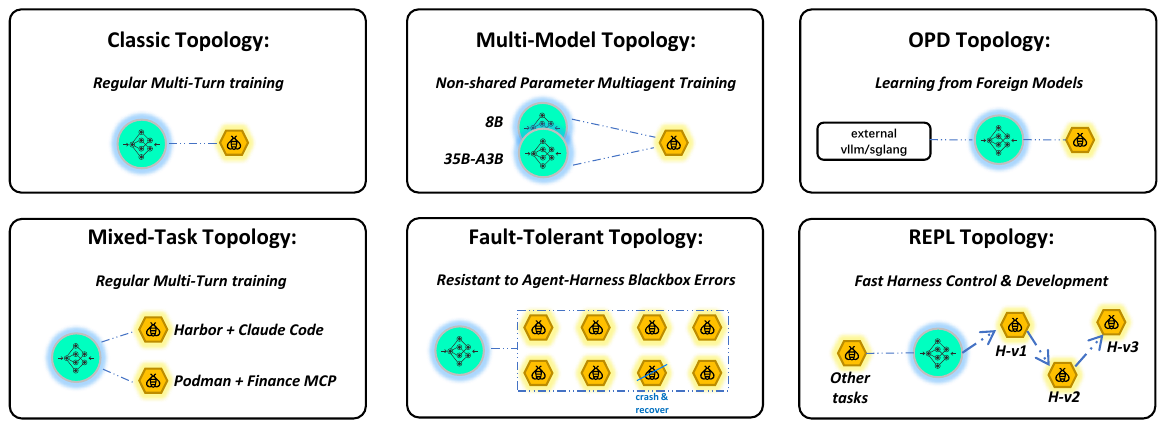}}
\caption{AgentJet topology variants supported by the swarm abstraction. The same serving-level interface covers classic single-model training, heterogeneous multi-model training, OPD-style learning from external models, mixed-task training, fault-tolerant execution, and REPL-style harness development.}
\label{fig:swarm_topology}
\end{figure}

\begin{table}[htbp]
\centering
\caption{Capabilities derived from the swarm decoupling, the design mechanism that produces each one, and the experiment that validates it.}
\label{tab:capabilities}
\resizebox{\textwidth}{!}{%
\begin{tabular}{@{}p{4.5cm} p{7.0cm} p{5.5cm}@{}}
\toprule
\textbf{Capability} & \textbf{Mechanism} & \textbf{Examples} \\
\midrule
Topology-configurable training & Reconfigure the number of trainable swarm servers and isolated swarm clients under the same serving interface (Sec.~\ref{sec:system_architecture}, Fig.~\ref{fig:swarm_topology}) & Shared-parameter, non-shared-parameter, mixed-task, fault-tolerant, and REPL-style training in this section \\
\addlinespace
Heterogeneous multi-model training & One trainable LLM per swarm server; per-episode routing to the correct policy (Sec.~\ref{sec:system_architecture}, Sec.~\ref{sec:interaction}) & Non-shared-parameter Werewolves and academic translation (Sec.~\ref{sec:adv-werewolves}) \\
\addlinespace
Mixed-task cocktail training & Incompatible runtimes isolated in separate clients; unified sample-pool interface (Sec.~\ref{sec:interaction}, Sec.~\ref{sec:pool_full}) & AppWorld + AIME joint training (Sec.~\ref{sec:cocktail-training}) \\
\addlinespace
Fault-tolerant training & Clients hot-swappable; accepted samples survive client failure (Sec.~\ref{sec:interaction}, Sec.~\ref{sec:repl}) & Crash recovery in AutoResearch campaigns (Sec.~\ref{sec:auto_research}) \\
\addlinespace
REPL-style debugging and evaluation & Fast-changing logic on the client, optimizer state on the server (Sec.~\ref{sec:repl}) & Framework-agnostic ablation, on-demand probes (Sec.~\ref{sec:experiments}) \\
\addlinespace
Decentralized training & CPU-only clients control remote GPU servers (Sec.~\ref{sec:system_architecture}, Sec.~\ref{sec:repl}) & Task onboarding from a CPU-only laptop (Sec.~\ref{sec:task_onboarding}) \\
\addlinespace
Automated research & Serving-layer separation exposes a stable training substrate to coding agents (Sec.~\ref{sec:repl}, Sec.~\ref{sec:auto_research}) & Onboarding and A3R pipelines (Sec.~\ref{sec:auto_research}) \\
\bottomrule
\end{tabular}%
}
\end{table}

Figure~\ref{fig:swarm_topology} summarizes the topology variants evaluated in this section.
The experiments follow a progression over the two axes of the swarm: the number of trainable servers and the number of isolated clients.
The \textbf{classic topology} fixes both axes at one server and one client, which is sufficient for shared-parameter multi-agent RL and ordinary multi-turn training.
The \textbf{multi-model topology} expands the server side of the swarm and removes the single-policy constraint inherent to conventional RL trainers, under which all agents in a team must share one model and one parameter set.
AgentJet instead places each trainable LLM on a separate swarm server; the client routes each agent action to the corresponding model, collects role-specific trajectories, and updates each policy exclusively from the experience it generated.

The \textbf{OPD} and \textbf{mixed-task} panels expand the client and data-source side of the same abstraction.
OPD-style training can treat external vLLM/SGLang deployments as foreign models that provide teacher behavior through the same request path, while AgentJet trains the student model behind the swarm server.
Mixed-task training addresses a more practical systems problem: agentic tasks often depend on mutually incompatible runtimes, such as sandboxes, proxies, MCP services, and reward services.
Because each unstable runtime is confined to its own client or container, a single invalid API key or sandbox crash remains local to that client, while the server trains on a mixed task distribution assembled through a unified sample interface.
The AppWorld--AIME experiment evaluates this topology by comparing one mixed run with separate single-task specialists.

The remaining panels show operational advantages that follow from the same state partitioning.
The \textbf{fault-tolerant topology} assigns everything expensive to reconstruct to the server and everything cheap to revise to the clients: a failed client can be restarted, terminated, or ignored without the server discarding optimizer state or already-accepted samples.
The \textbf{REPL topology} uses this fault tolerance for iterative development and evaluation: a researcher can edit an agent loop or reward function, soft-restart the client within seconds, or attach a fresh evaluation client to an already-running server without reloading the model.
Because clients are CPU-only processes that issue routing credentials and inference requests, this also yields Tinker-like~\citep{tinker} \textbf{decentralized training}: a GPU-less laptop can control a swarm that trains a multi-agent system built from heterogeneous LLMs, bounded only by available compute and networking.
Finally, the same REPL substrate extends upward into \textbf{automated research}: 
clients carry no irreplaceable state and reattach in seconds, so a coding agent can generate clients, implement rewards, launch rollouts, inspect logs, and recover from failures without touching the server-side optimizer.
At larger scale, such agents can orchestrate multi-day, multi-server experiment campaigns over the same serving-layer training interface.
The following subsections instantiate this progression: classic shared-parameter training, multi-model non-shared-parameter training, mixed-task training over isolated runtimes, and regular multi-turn training that stresses the efficiency and operational benefits of the same topology.

\subsection{Shared-Parameter Multi-Agent Training}

We begin with the classic swarm topology in Fig.~\ref{fig:swarm_topology}, where one swarm server is paired with one swarm client.
This topology corresponds to shared-parameter multi-agent reinforcement learning (MARL): all trainable agents share a common model hosted by the server, while the client executes the multi-agent environment and assigns rewards.
Each trainable agent still acts independently from its own observation, and static (non-trainable) LLM agents can coexist in the same client-side environment as opponents, collaborators, reward judges, or subagents.
Thus, the experiment employs a \textbf{1-server-1-client} swarm RL network in which a single server hosts both the shared-parameter LLM and its optimizer.

\subsubsection{Werewolves RPG Reinforcement Learning}
\label{sec:shared-werewolves}

The Werewolves game is a social deduction role-playing game that serves as a challenging testbed for multi-agent reinforcement learning.
It presents a Partially Observable Markov Decision Process (POMDP)~\citep{kaelbling1998pomdp} where agents must make decisions based on incomplete information
while engaging in complex social interactions, including deception, coalition formation, and strategic voting.

We configure a 9-player game $(n_{\text{ww}}, n_{\text{vl}}, n_{\text{sr}}, n_{\text{wt}}, n_{\text{ht}}) = (3, 3, 1, 1, 1)$,
denoting the number of werewolves, villagers, seers, witches, and hunters respectively.
Werewolves secretly kill players at night while hiding their identity;
villagers have no special abilities; the seer can check the identity of one player each night;
the witch has one healing potion and one poison potion; and the hunter can shoot a player upon death.
The trainable agents use shared-parameter \textbf{Qwen2 model (7B or 14B)},
while opponents are controlled by a static \textbf{Qwen3-235B-A22B} model as a strong adversary.
This creates an asymmetric training scenario where lighter models learn to compete against significantly larger opponents.

The topology remains minimal because all trainable roles update the same policy and the Werewolves environment does not require multiple incompatible runtimes.
The single swarm client therefore runs the complete game, routes trainable-agent calls to the shared server endpoint, and assigns a sparse, turn-level reward based on the game outcome:
$r=1$ if the trainable faction wins, $r=0$ otherwise.
An exception penalty of $r=-0.1$ is applied when game execution fails due to illegal or unexpected agent actions.
The seat numbers of all agents are randomly assigned at the start of each episode to prevent overfitting to specific roles or positions.

\begin{table}[t]
\centering
\caption{Werewolf training experiments with different trainable agent configurations. Abbreviations: \roleMain{ww}=werewolf, \roleMain{vl}=villager, \roleMain{sr}=seer, \roleMain{wt}=witch, \roleMain{ht}=hunter, SR=success rate.}
\label{tab:werewolves_results}
\resizebox{\textwidth}{!}{%
\begin{tabular}{clclcc}
\toprule
\textbf{Exp} & \textbf{Trainable} & \textbf{Size} & \textbf{Static Model (Qwen3-235B-A22B)} & \textbf{Initial SR} & \textbf{Final SR} \\
\midrule
1 & \roleInTable{ww} & 7B & opponents: \{\roleInTable{vl}, \roleInTable{sr}, \roleInTable{wt}, \roleInTable{ht}\} & 23.0 & 47.2 \\
2 & \roleInTable{ww} & 14B & opponents: \{\roleInTable{vl}, \roleInTable{sr}, \roleInTable{wt}, \roleInTable{ht}\} & 40.9 & 64.7 \\
3 & \roleInTable{sr} & 14B & opponents: \{\roleInTable{ww}\}, collaborators: \{\roleInTable{vl}, \roleInTable{wt}, \roleInTable{ht}\} & 38.5 & 46.5 \\
4 & \roleInTable{wt} & 14B & opponents: \{\roleInTable{ww}\}, collaborators: \{\roleInTable{vl}, \roleInTable{sr}, \roleInTable{ht}\} & 38.8 & 38.9 \\
5 & \roleInTable{ht} & 14B & opponents: \{\roleInTable{ww}\}, collaborators: \{\roleInTable{vl}, \roleInTable{sr}, \roleInTable{wt}\} & 31.9 & 34.5 \\
6 & \roleInTable{sr}, \roleInTable{wt}, \roleInTable{ht} & 14B & opponents: \{\roleInTable{ww}\}, collaborators: \{\roleInTable{vl}\} & 22.9 & 35.9 \\
7 & \roleInTable{vl}, \roleInTable{sr}, \roleInTable{wt}, \roleInTable{ht} & 14B & opponents: \{\roleInTable{ww}\} & 23.9 & 41.6 \\
\bottomrule
\end{tabular}%
}
\end{table}

\paragraph{Results and Analysis.}
Table~\ref{tab:werewolves_results} reports success rates across seven shared-parameter configurations covering both factions.
Training the werewolf faction yields the largest absolute gains:
the 7B model improves from 23.0\% to 47.2\% (Exp~1), and the 14B model from 40.9\% to 64.7\% (Exp~2),
despite both facing a static 235B opponent ensemble.
Within the villager faction, training a single specialized role produces the highest per-role win rate when that role is the seer (Exp~3, 38.5\%~$\to$~46.5\%),
while the witch (Exp~4) and hunter (Exp~5) show only marginal improvement,
suggesting their narrower action spaces and one-shot abilities offer fewer learnable decision points.
Jointly training the three special roles (Exp~6, 22.9\%~$\to$~35.9\%) and the full non-werewolf team (Exp~7, 23.9\%~$\to$~41.6\%) recovers most of the gains
and confirms that shared-parameter training in AgentJet scales to heterogeneous role sets without per-role reward shaping.
Beyond these quantitative gains, the trained agents also exhibit qualitative behavioral improvements, which we examine through concrete trajectories below.

\begin{figure}[t]
\centering
\includegraphics[width=0.85\linewidth]{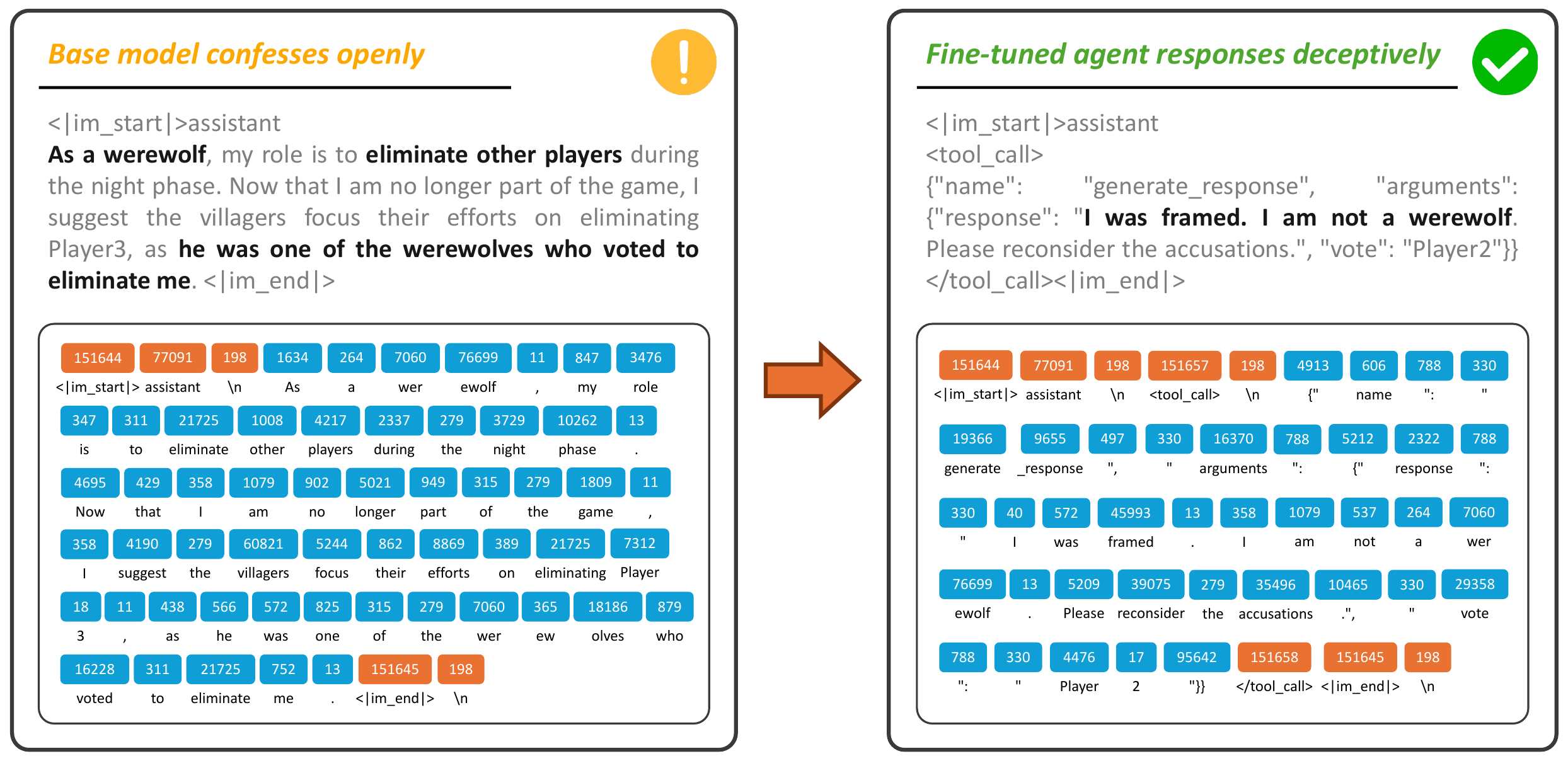}
\caption{Token-level visualization rendered by the AgentJet Beast-Logger of werewolf behavior before and after fine-tuning. Left: once voted out, the base model openly confesses. Right: the fine-tuned agent instead emits a deceptive \texttt{generate\_response}.}
\label{fig:werewolf_case_deception}
\end{figure}

\paragraph{Case Study.}
A significant role-playing improvement is observed over the course of training, visible in concrete game trajectories:
\begin{itemize}
    \item \textbf{Role-playing consistency}: When voted out, the original model tends to reveal its identity as a \roleMain{ww}, whereas after fine-tuning the agent maintains its cover, continuing to deceive its opponents and protect its teammates. Figure~\ref{fig:werewolf_case_deception} contrasts the two behaviors at the token level, as rendered by the AgentJet logger.
    \item \textbf{Social deception strategies}: The agent develops multiple tactics for winning, including misdirection (``Let us keep an eye on the seer and the witch. They could be werewolves trying to hide.''), appeals to reason (``We need to be wary of fake seers and watch for inconsistencies in stories; Player-Y as hunter should act carefully.''), and taking advantage of the suspicion between non-werewolf players to eliminate opponents.
    \item \textbf{Implicit coordination}: Werewolf agents learn to coordinate voting without explicit communication, leveraging shared-parameter training to develop emergent team strategies.
\end{itemize}

\subsection{Non-Shared Parameter Multi-Agent Training}

We next expand the server side of the classic topology while keeping the client-side environment fixed.
Whereas shared-parameter training is appropriate for homogeneous agent teams, many real-world settings require heterogeneous agents with distinct objectives, capabilities, or adversarial relationships.
Non-shared-parameter multi-agent training gives each agent, or each agent group, independent model weights, enabling the study of emergent behavior in competitive, cooperative, and mixed-motive environments.
AgentJet implements this multi-model topology by assigning each trainable policy to a dedicated swarm server, while a single swarm client runs the shared environment and routes each agent action to the correct server.
In these experiments, we adopt a \textbf{many-server-1-client} swarm network configuration.

\subsubsection{Adversarial Werewolves RPG Reinforcement Learning}
\label{sec:adv-werewolves}

Section~\ref{sec:shared-werewolves} used the classic topology: all trainable Werewolves roles updated one shared model through one server.
The multi-model topology changes only the server allocation, not the game environment.
A single Werewolves client still manages the episode, static opponents, and reward, but different trainable role groups now route to different swarm servers and update independent policies.
This isolates role-specific experience while preserving interaction inside the same social game.
Here, we focus on \textbf{cooperation}, where all trainable models remain in the same faction.

\begin{table}[t]
\centering
\caption{Non-shared parameter multi-agent training experiments. Multiple 14B models with LoRA are trained simultaneously, each controlling different roles. Abbreviations: \roleMain{ww}=werewolf, \roleMain{vl}=villager, \roleMain{sr}=seer, \roleMain{wt}=witch, \roleMain{ht}=hunter, SR=success rate.}
\label{tab:werewolves_multiagent}
\resizebox{\textwidth}{!}{%
\begin{tabular}{clllcc}
\toprule
\textbf{Exp} & \textbf{Trainable Model} & \textbf{Trained Role} & \textbf{Static Model (235B)} & \textbf{Initial SR} & \textbf{Final SR} \\
\midrule
\multirow{2}{*}{1} & M1 (14B-LoRA, 4 GPUs) & \roleInTable{ht}, \roleInTable{vl} & \multirow{2}{*}{opponents: \{\roleInTable{ww}\}} & \multirow{2}{*}{22.5} & \multirow{2}{*}{30.5} \\
                   & M2 (14B-LoRA, 4 GPUs) & \roleInTable{wt}, \roleInTable{sr} & & & \\
\midrule
\multirow{3}{*}{2} & M1 (14B-LoRA, 3 GPUs) & \roleInTable{vl} & \multirow{3}{*}{opponents: \{\roleInTable{ww}\}} & \multirow{3}{*}{22.0} & \multirow{3}{*}{34.2} \\
                   & M2 (14B-LoRA, 3 GPUs) & \roleInTable{sr}, \roleInTable{wt} & & & \\
                   & M3 (14B-LoRA, 2 GPUs) & \roleInTable{ht} & & & \\
\midrule
\multirow{3}{*}{3} & M1 (14B-LoRA, 3 GPUs) & \roleInTable{ww-1} & \multirow{3}{*}{opponents: \{\roleInTable{vl}, \roleInTable{sr}, \roleInTable{wt}, \roleInTable{ht}\}} & \multirow{3}{*}{40.8} & \multirow{3}{*}{66.5} \\
                   & M2 (14B-LoRA, 3 GPUs) & \roleInTable{ww-2} & & & \\
                   & M3 (14B-LoRA, 2 GPUs) & \roleInTable{ww-3} & & & \\
\midrule
\multirow{2}{*}{4} & M1 (14B-LoRA, 4 GPUs) & 50\% random non-\roleInTable{ww} & \multirow{2}{*}{opponents: \{\roleInTable{ww}\}} & \multirow{2}{*}{24.0} & \multirow{2}{*}{37.0} \\
                   & M2 (14B-LoRA, 4 GPUs) & remaining 50\% non-\roleInTable{ww} & & & \\
\bottomrule
\end{tabular}%
}
\end{table}

Table~\ref{tab:werewolves_multiagent} presents results from four non-shared parameter training configurations.
Several findings emerge:

In general, werewolves are easier to train than the villager faction. This is consistent with Table~\ref{tab:werewolves_results}, where shared-parameter training of werewolves achieves a higher final success rate (64.7\%).

An interesting finding emerges when comparing shared-parameter and non-shared-parameter training for werewolves.
Table~\ref{tab:werewolves_results} Exp~2 uses a single 14B model for all werewolves, achieving 64.7\% final success rate,
while Table~\ref{tab:werewolves_multiagent} Exp~3 uses three separate 14B-LoRA models, achieving 66.5\%.
Despite nearly identical initial success rates (40.9\% vs. 40.8\%),
training separate models for each werewolf player yields a 1.8\% improvement.
This gain comes from behavioral diversity:
when werewolves share parameters, they tend to exhibit similar speech patterns and voting behaviors,
which experienced villagers can exploit to identify the werewolf team.
With independent parameters, each werewolf develops a distinct persona,
breaking this correlation and making coordinated identification significantly harder.
This result highlights the unique advantage of non-shared parameter training in social deduction games,
where behavioral diversity is crucial for deception.

For the villager faction, we observe nuanced trade-offs between specialization and generalization.
Comparing Exp~1 (2 models) and Exp~2 (3 models),
finer-grained role specialization yields better performance (34.2\% vs 30.5\%),
as the hunter agent can develop strategies for its unique night-kill ability without compromising other roles.
However, Exp~4 with random role assignment achieves the best result (37.0\%),
outperforming both fixed-assignment configurations.
This suggests that effective cooperation requires perspective-taking:
by experiencing the constraints and decision-making processes of other collaborator roles,
each model develops a better understanding of teammate behaviors,
enabling more effective coordination during the cooperation process.

\subsubsection{Hierarchical Multi-agent Academic Translation}
\label{sec:academic-translation}

The same topology also applies to hierarchical agent workflows in which different stages benefit from different model capacities.
Academic translation at scale demands high throughput, low latency, and precise adherence to domain-specific conventions.
These requirements favor smaller models that can process massive corpora efficiently,
yet small models typically struggle with instruction following in long contexts.
A natural solution is to keep the high-throughput translation and revision stages on a smaller trainable model, while routing the harder terminology-review stage to a stronger model endpoint.
AgentJet expresses this as the same client-controlled routing problem: the translation client executes the proposal--review--revision workflow, and the trainable stages contribute trajectories to the corresponding swarm server for reinforcement learning.

The pipeline consists of three agents collaborating to translate English academic abstracts into Chinese.
Agent~1 (Qwen2.5-7B-Instruct~\citep{qwen2.5}) produces an initial rough translation,
focusing on fluency and adherence to Chinese academic writing conventions:
replacing first-person pronouns (e.g., ``we'') with impersonal constructions (e.g., ``this study''),
intelligently handling abbreviations (using Chinese for short terms, retaining English for long ones with full expansion on first mention),
and adjusting word order to match Chinese rhetorical emphasis.
Agent~2 (Qwen2.5-14B-Instruct) reviews the rough translation and detects errors in discipline-specific proper nouns,
outputting a structured JSON list of corrections with original term, erroneous translation, error reason, and suggested fix.
Agent~3 (the same 7B model) applies these corrections to produce the final polished translation.
This hierarchical design leverages model specialization:
the larger 14B model handles the cognitively demanding terminology verification,
while the efficient 7B model handles the bulk translation work.

The reward model is constructed using OpenJudge, an LLM-as-a-judge~\citep{zheng2023llmjudge} grader, with a rigorous evaluation protocol.
Rather than assessing general translation quality,
the grader focuses on specific error categories demonstrated through few-shot examples:
first-person pronoun misuse, abbreviation translation errors, word order problems violating Chinese academic style,
subject-verb inconsistencies from improper restructuring, inappropriate colloquial word choices,
redundant punctuation disrupting reading flow, and unclear or missing subjects.
Translations are scored on a 0--2 scale:
0 for severe errors impairing readability, 1 for noticeable errors reducing reading efficiency, and 2 for error-free output.

Table~\ref{tab:translation_examples} presents representative examples comparing the fine-tuned 7B model against baselines.
The base 7B model exhibits characteristic failures:
mixed Chinese-English output, loss of semantic details, and retention of inappropriate first-person pronouns.
After fine-tuning, the model correctly expands abbreviations on first mention,
replaces ``we'' with appropriate impersonal subjects,
and accurately translates domain-specific terms.

\begin{table*}[htbp]
\centering
\caption{Translation comparison between base and fine-tuned Qwen2.5-7B-Instruct.
The fine-tuned model correctly handles abbreviation expansion, first-person pronoun replacement, and proper noun translation.}
\label{tab:translation_examples}
\includegraphics[width=\textwidth]{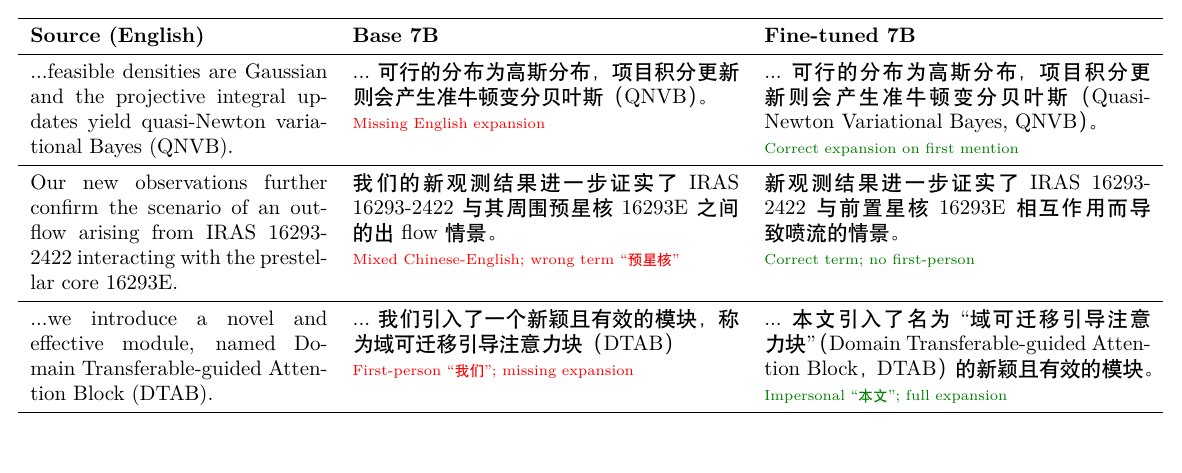}
\end{table*}

\subsection{Multi-Task Training}
\label{sec:cocktail-training}

After varying the number of trainable servers, we vary the number of client-side runtimes.
Mixed-task training uses one trainable swarm server but attaches multiple isolated swarm clients, each running a different task environment and reward pipeline.
This topology tests whether a single trainable model can be jointly optimized over a heterogeneous mixture of agentic tasks, a setting we call \emph{cocktail training}.
Each task brings its own runtime: AppWorld requires a containerized execution backend with stateful application APIs, while AIME mathematical reasoning only needs a lightweight verifier.
In a monolithic trainer, the two stacks would have to coexist inside the same process, causing dependency conflicts and unstable rollouts.
AgentJet avoids this issue: the AppWorld and AIME swarm clients run in separate environments and stream rollouts to a shared swarm server, which performs unified policy gradient updates over the mixed batch.

In this experiment, we train Qwen3-8B with GRPO on a single 8-GPU node.
The cocktail run instantiates the mixed-task topology by attaching two swarm clients, AppWorld and AIME, to one swarm server, with a per-step batch of 16 AppWorld trajectories plus 16 AIME trajectories and GRPO group size 8.
We compare it against two separate single-task runs, AppWorld only and AIME only, that use a per-step batch of 32 trajectories and otherwise share the same model, optimizer, and hardware.

\begin{figure}[t]
\centering
\includegraphics[width=0.98\linewidth]{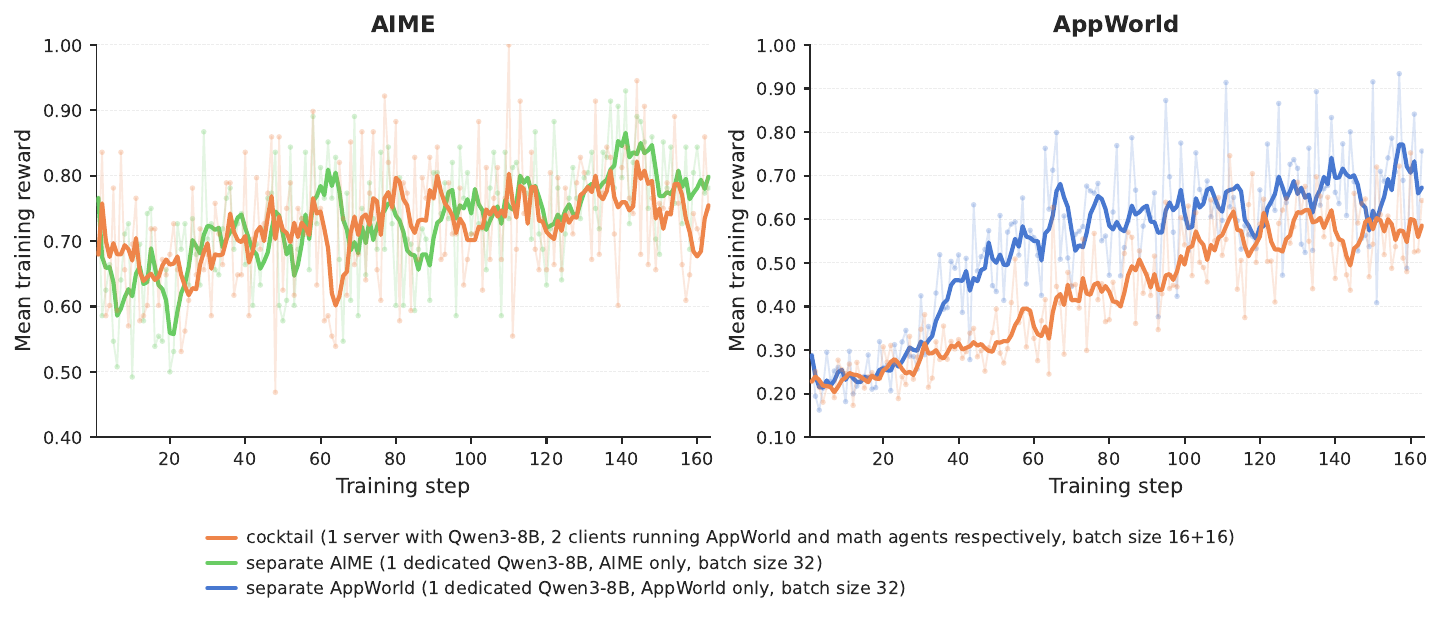}
\caption{Per-step training reward on AIME (left) and AppWorld (right), comparing cocktail joint training (orange, the same single run shown in both panels) against two separate single-task specialists: a dedicated AIME run (green) and a dedicated AppWorld run (blue). Thin lines and markers show raw step rewards; thick lines show a 5-step rolling average. Both regimes use Qwen3-8B with GRPO on the same 8-GPU node; the cocktail run uses a per-step batch of 16 (AppWorld) plus 16 (AIME), while each separate run uses a per-step batch of 32. All curves are clipped to the common 163-step range covered by every run.}
\label{fig:cocktail_vs_separate}
\end{figure}

Figure~\ref{fig:cocktail_vs_separate} summarizes the result.
On AIME, the two regimes track each other closely: their reward curves overlap for most of training and their means over the full run differ by less than one reward point (0.72 versus 0.73), with the separate run opening only a small late-stage margin (a last-20-step mean of 0.80 versus 0.75).
On AppWorld, the gap is more pronounced: the separate run holds a consistent margin over the cocktail run, roughly 10 reward points higher both on average over training and over the last 20 steps (0.68 versus 0.58).
This asymmetric tax is consistent with the interpretation that AppWorld requires longer, tool-using trajectories whose gradients are more easily diluted when mixed with shorter, single-shot AIME rollouts.
We do not claim that cocktail training matches separate training on every task. On a tool-heavy benchmark like AppWorld a dedicated specialist run remains the stronger choice, and the curves should be read with that asymmetry in mind.

The value of cocktail training lies elsewhere.
First, it produces \emph{one} model that is competent across the entire task mixture rather than a separate specialist per benchmark. Because the model is exposed within a single optimization to qualitatively different reasoning regimes (long-horizon tool use in AppWorld, short-horizon symbolic reasoning in AIME), it must maintain a shared representation that supports both, which we expect to translate into better cross-task generalization than a specialist that has only ever seen one regime.
Second, it substantially reduces overall training cost.
Producing $N$ specialists requires $N$ independent training campaigns, each paying its own warmup, optimizer state, and rollout-cluster setup; cocktail training amortizes all of these across one shared run on one shared model copy. Adding a task increases per-step rollout volume but does not multiply the run count, the GPU-time footprint, or the engineering overhead.
These results suggest that cocktail training is most appropriate when the goal is a single deployable generalist at low marginal cost per added task, whereas separate single-task training remains preferable when the explicit objective is to maximize performance on one benchmark.

A common recipe for building a single multi-skill model is the two-stage \emph{on-policy distillation} (OPD)~\citep{agarwal2024opd} pipeline: first, train a separate specialist model for each task with RL; then, distill all specialists into one student by sampling trajectories on-policy from the student and matching the per-token distribution of each specialist on its respective task.
OPD is effective but expensive: it requires $N+1$ training runs for $N$ tasks, $N$ teacher checkpoints kept resident for sampling, and a careful gating mechanism to route each student rollout to the right teacher.
Cocktail training, as enabled by the AgentJet swarm, offers an alternative that collapses this pipeline into a single run.
Instead of distilling skills back together after the fact, the mixed-task topology lets the swarm server receive on-policy rollouts from every task client simultaneously and apply a unified policy gradient update over the mixed batch.
There is no teacher, no second stage, and no routing logic; the only ``mixing'' decision is the relative rollout budget per client, which is controlled directly by the per-client batch configuration.
Crucially, cocktail training optimizes for the same goal as OPD: a single multi-skill model. The trade-off is different, however. Cocktail training accepts a per-task gap relative to dedicated specialists in exchange for skipping the specialist-training and distillation stages entirely, removing $N$ teacher checkpoints and the associated routing logic from the pipeline.
We view the two approaches as complementary: OPD remains attractive when strong specialists already exist, or when tasks demand teacher supervision beyond reward signals, while cocktail training is preferable when one wishes to obtain a multi-skill generalist from scratch in a single training campaign with bounded total compute.

\subsection{Regular Multi-Turn Training}

The final group of experiments returns to the classic one-server, one-client topology and stresses a different dimension of the swarm abstraction: long multi-turn interaction.
Here, the topology is not expanded by adding more trainable models or more task clients.
Instead, the same server--client boundary must preserve complete multi-turn trajectories, avoid redundant actor-update computation, tolerate changes in client-side agent code, and support tool-intensive workflows whose runtimes are noisy.
These experiments therefore connect the classic, fault-tolerant, and REPL topology panels in Fig.~\ref{fig:swarm_topology}.

\begin{figure}[t]
\centering
\includegraphics[width=0.98\linewidth]{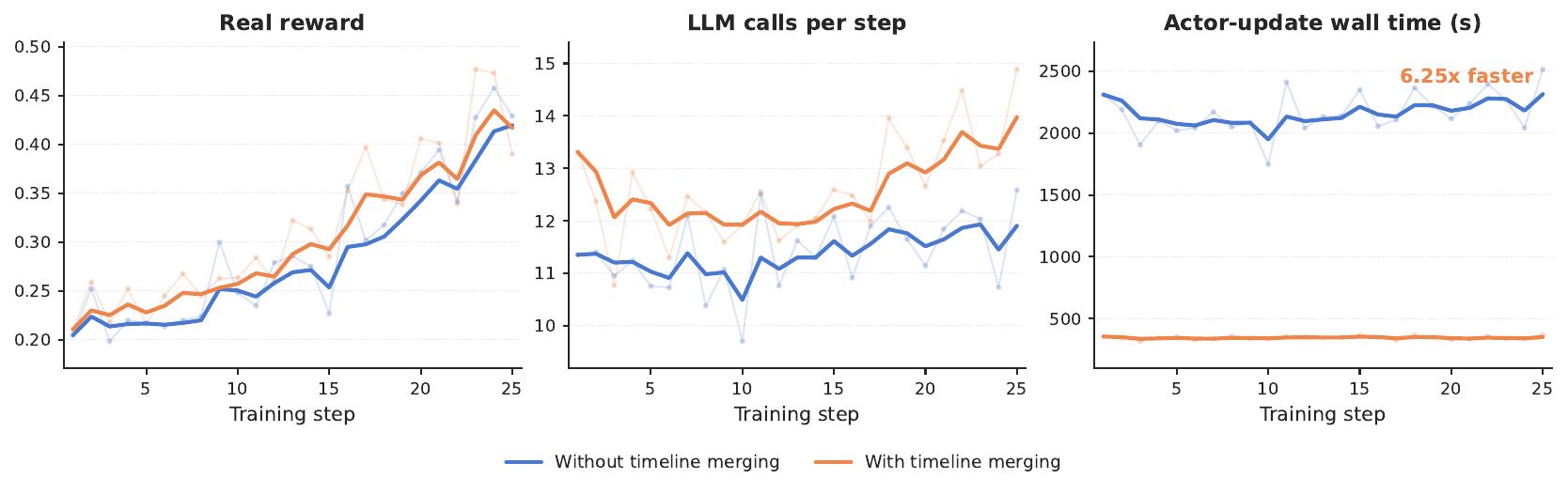}
\caption{Effect of timeline merging on a multi-turn AppWorld swarm RL task
(first 25 training steps; two runs identical except for the merging switch).
Left: real reward; center: LLM calls per step; right: actor-update
wall-clock time. The reward and call-count curves track each other closely
while the actor-update time is reduced by 6.25$\times$ on average, so the
large speedup comes at no cost to training quality.}
\label{fig:timeline_merging}
\end{figure}

\subsubsection{Fully Automated Timeline Merging}
AgentJet automatically consolidates the redundant context that accumulates
across multi-turn agent rollouts (Section~\ref{sec:timeline_merging}),
requiring no change to the task, model, learning algorithm, or client-side agent loop.
At this topology level, timeline merging is a server-side optimization enabled by the fact that the swarm server observes all routed LLM calls for an episode while the client remains responsible only for executing the agent environment.
Figure~\ref{fig:timeline_merging} compares two otherwise identical AppWorld swarm RL experiments over the first 25 training steps, one with timeline merging disabled and one with it enabled.
AppWorld~\citep{trivedi2024appworld} is an interactive coding benchmark in which an agent completes everyday digital tasks (e.g., managing email and music) by issuing multi-turn API calls against a high-fidelity simulated world of nine apps and roughly 450 APIs.
The real-reward curves overlap closely and the per-step LLM-call counts stay comparable ($11.4 \pm 0.7$ versus $12.6 \pm 1.0$), confirming that merging preserves training quality and agent behavior;
meanwhile, the average per-step actor-update wall time falls from $2160 \pm 171$\,s to $346 \pm 13$\,s, a 6.25$\times$ speedup.

\subsubsection{Performance Stability and Consistency}
\begin{figure}[t]
\centering
\includegraphics[width=0.98\linewidth]{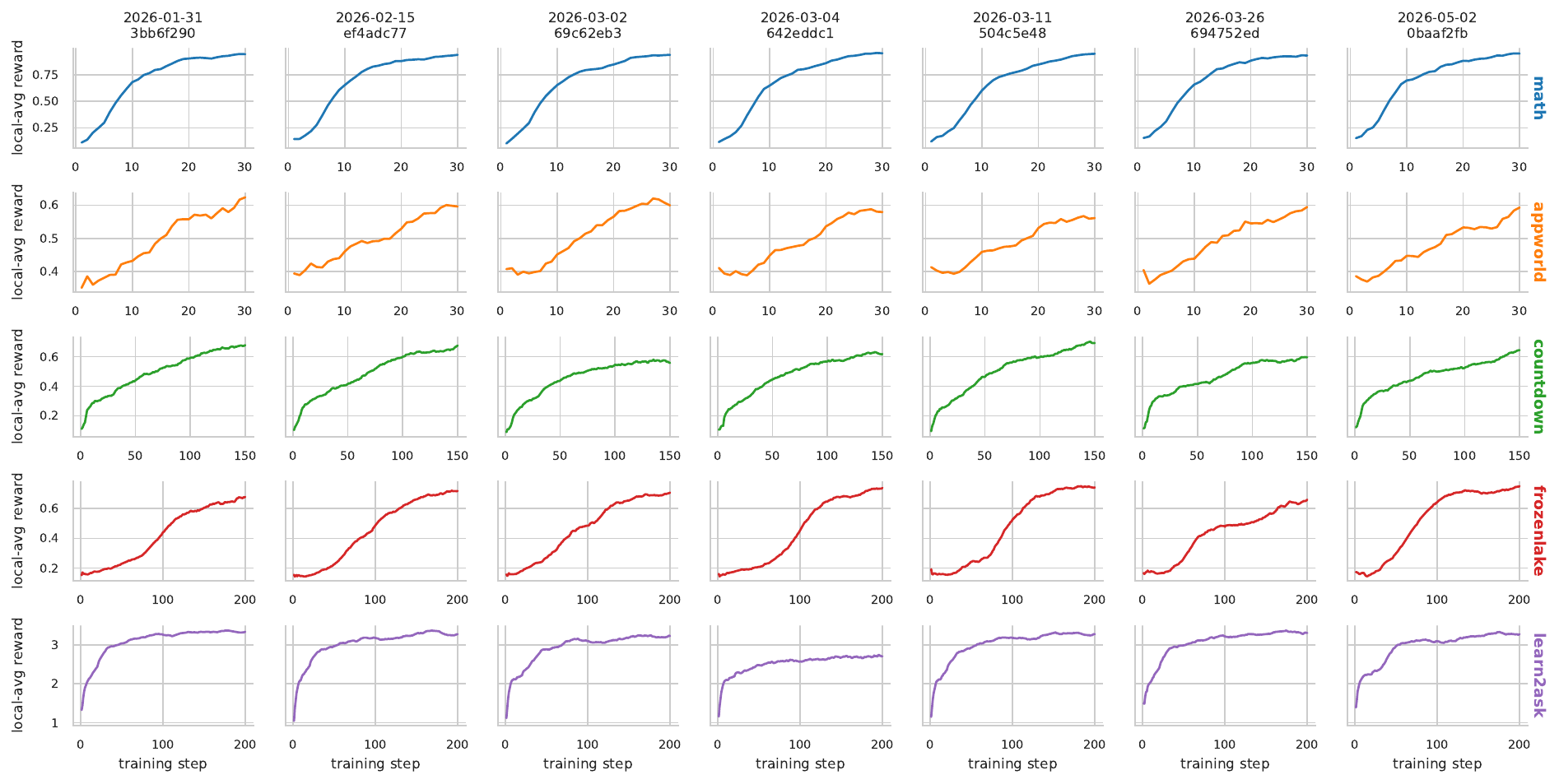}
\caption{Per-step rolling-average reward across benchmark tasks (rows) and
AgentJet git commits (columns, sorted by commit date). These baselines are performed regularly by auto research agents, reflecting the stability of the AgentJet's framework during the rapid iteration.}
\label{fig:benchmark_stability}
\end{figure}

As shown in Figure~\ref{fig:benchmark_stability}, the continuous-benchmarking
experiment\footnote{\url{https://benchmark.agentjet.top}} re-runs a fixed suite of multi-turn RL tasks regularly for important versions.
This experiment exercises the fault-tolerant topology at the software-evolution scale: client-side tasks, rewards, and harness code are repeatedly relaunched against the same serving-level training abstraction while the server-side optimizer path remains stable.
Each row in Figure~\ref{fig:benchmark_stability} is one benchmark task and
each column is one git commit (ordered by commit date); the per-step training
reward curves overlap closely, confirming that algorithmic refactors and infrastructure optimizations do not regress training quality.

\subsubsection{Framework Agnostic Agent Loops}
\begin{figure}[t]
\centering
\includegraphics[width=0.98\linewidth]{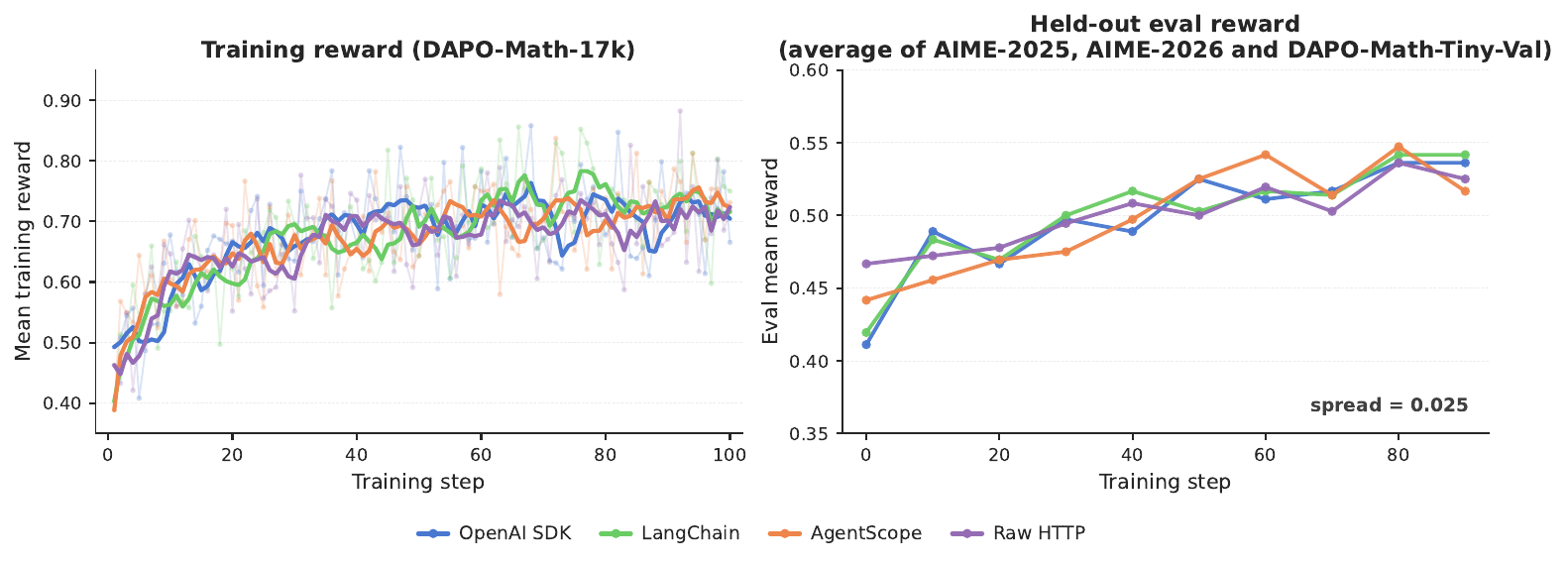}
\caption{Framework-agnostic training of four agent-loop implementations
(OpenAI SDK, LangChain, AgentScope, Raw HTTP) under an otherwise identical
Qwen3-8B GRPO recipe on DAPO-Math-17k for 100 steps.
Left: per-step training reward (\texttt{critic/rewards/mean}); thin lines
and markers are raw values, thick lines are a 5-step rolling average.
Right: held-out evaluation reward averaged across AIME-2025, AIME-2026, and
DAPO-Math-Tiny-Val, sampled at ten checkpoints; the final-step
cross-arm spread is $0.025$, and the maximum cross-arm gap at any aligned
checkpoint stays below $0.04$, confirming that the choice of agent-loop
framework is transparent to AgentJet's training.}
\label{fig:framework_ablation}
\end{figure}

AgentJet captures the RL training signal at the swarm's OpenAI-compatible endpoint
rather than inside the agent loop.
At the framework level, this experiment instantiates the REPL topology: different clients can be attached, replaced, or reimplemented without changing the server-side optimizer, as long as they issue OpenAI-style \texttt{chat.completions} requests against the endpoint.
Those requests are captured by the per-episode timeline cache, and the same dense-LLM update follows.
We verify this empirically with a four-arm framework ablation in which the same
multi-turn math-reasoning GRPO recipe (Qwen3-8B base, DAPO-Math-17k~\citep{yu2025dapo} training data
with a Python tool, validation on AIME-2025, AIME-2026, and DAPO-Math-Tiny-Val,
100 training steps, $\text{batch}=32$, $\text{grpo\_repeat}=8$, $\text{val\_pass\_n}=4$,
default learning rate) is driven by four different agent-loop implementations:
the OpenAI Python SDK, LangChain, AgentScope, and a hand-written Raw HTTP client.
The system prompt, Python tool sandbox, reward function, dataset, hyperparameters,
and per-step batch/group structure are held byte-identical across the four runs;
the only variable is the agent-loop framework.
As shown in Figure~\ref{fig:framework_ablation}, the four training and evaluation
reward curves overlap closely throughout training. The final-checkpoint
(step~90) evaluation rewards averaged across the three held-out sets are
$0.536$ (OpenAI SDK), $0.542$ (LangChain), $0.517$ (AgentScope), and $0.525$ (Raw HTTP);
the spread across the four frameworks is only $0.025$, and the maximum cross-arm
gap at any aligned checkpoint stays below $0.04$ for the entire run.
This confirms that AgentJet's training capture is framework-agnostic:
practitioners can pick whichever agent framework best fits their workflow
and obtain essentially the same RL training dynamics.

\subsubsection{Financial Deep Research Agent Training}
\label{sec:deepfinance}

This study uses AgentJet to train a financial deep research agent through reinforcement learning, extending the same topology to a longer and noisier tool-using workload.
The agent receives open-ended analytical queries, autonomously plans multi-round information retrieval through financial tools, and produces structured, citation-grounded research reports.
Unlike the bounded episodes in previous experiments, financial deep research episodes involve 10+ turns of tool interaction, real-world API noise (timeouts, rate limits, non-deterministic responses), and require balancing multiple conflicting objectives: evidence traceability, analytical depth, and presentation quality.
The experiment therefore tests whether the server can preserve long-horizon training state while the client mediates volatile external tools and reward computation.

\paragraph{Agent workflow.}
The agent follows a two-phase research protocol.
In the first phase, the agent outputs a structured research outline with key questions for each section. No tool calls are made.
In the second phase, the agent executes iterative retrieval rounds (up to 10 turns), calling up to 3 tools per turn from a suite of 19 financial tools exposed through Finance-MCP\footnote{\url{https://github.com/flowllm-ai/finance-mcp}} via the Model Context Protocol~\citep{mcp2024}.
Tools cover entity extraction, stock price analysis, web search, and structured data crawling from financial data providers.
After sufficient evidence is gathered, the agent generates a Markdown report with numbered citation markers \texttt{[n]} linked to a references section.
An EnvService layer mediates all tool calls and implements MongoDB-backed caching: identical \texttt{(tool, args)} pairs execute once and results are reused across GRPO rollouts, reducing cost and stabilizing reward variance against external API volatility.

\paragraph{Reward design.}
We decompose the reward into four scoring dimensions plus a tool-call penalty:
\begin{equation}
    r = 0.5 \cdot r_{\text{rm}} + 0.2 \cdot r_{\text{present}} + 0.1 \cdot r_{\text{ground}} + 0.2 \cdot r_{\text{audit}} + r_{\text{penalty}},
\end{equation}
where $r_{\text{rm}}$ evaluates analytical sufficiency via pairwise comparison against expert references, $r_{\text{present}}$ scores presentation quality across 8 sub-criteria (scannability, information structuring, editorial clarity), $r_{\text{ground}}$ measures citation coverage and authenticity, and $r_{\text{audit}}$ performs logical entailment verification on each citation.
The tool-call penalty assigns $-1.0$ for zero tool calls, $-0.5$ for 1--2 calls, and $0$ for $\geq$3 calls, blocking degenerate no-retrieval strategies.
To stabilize the reward signal, each dimension uses a two-stage pipeline: a judge LLM extracts structured claims, evidence relationships, and verdicts, then deterministic rule-based logic computes scores from the extraction, avoiding the variance of end-to-end LLM scoring.

\paragraph{Training configuration.}
Training uses Qwen3-30B-A3B with GRPO (group size 4, batch size 32) over approximately 1{,}000 synthesized financial queries spanning macro analysis, industry research, event interpretation, stock analysis, and company research.

\begin{figure}[t]
\centering
\includegraphics[width=0.98\linewidth]{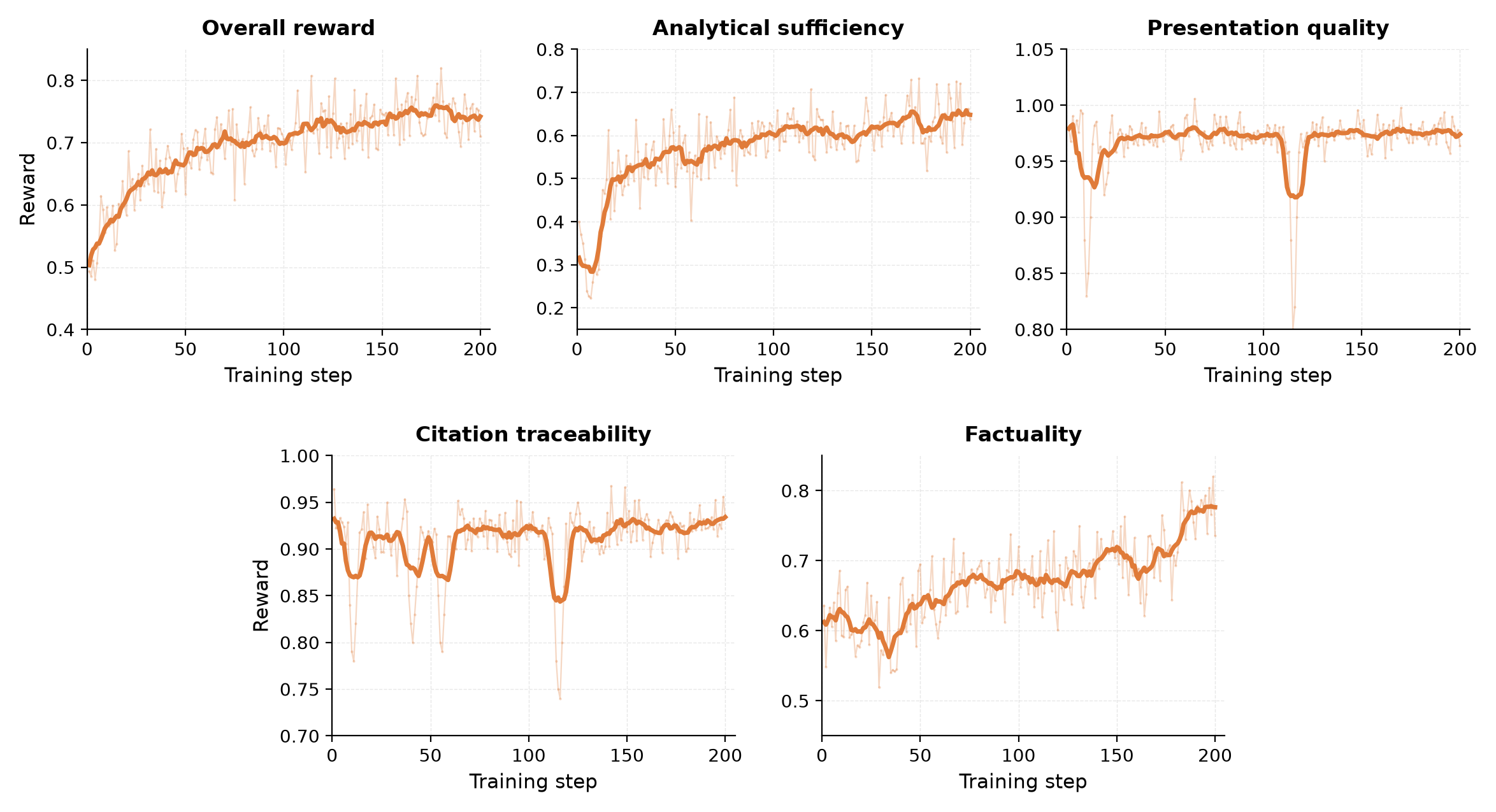}
\caption{Training dynamics over 200 GRPO steps.
Top: overall reward, analytical sufficiency $r_{\text{rm}}$, presentation quality.
Bottom: citation traceability $r_{\text{ground}}$, factuality $r_{\text{audit}}$.
Thin lines are raw per-step values; thick lines are 10-step rolling averages.}
\label{fig:deepfinance_training}
\end{figure}

\paragraph{Experimental results.}
Figure~\ref{fig:deepfinance_training} shows the training dynamics and Table~\ref{tab:deepfinance_results} summarizes external evaluation on DeepResearch Bench~\citep{du2025deepresearchbench}.

\begin{table}[t]
\centering
\caption{External evaluation on DeepResearch Bench. Improvements span both finance and non-finance subsets, suggesting the learned research process transfers beyond the financial training domain.}
\label{tab:deepfinance_results}
\small
\begin{tabular}{lccc}
\toprule
\textbf{Model} & \textbf{Finance} & \textbf{Others} & \textbf{Overall} \\
\midrule
Qwen3-30B-A3B-Instruct (base) & 0.184 & 0.118 & 0.127 \\
Tongyi DeepResearch & 0.296 & 0.274 & 0.277 \\
Claude 3.7 & 0.417 & 0.423 & 0.422 \\
\textbf{Ours (RL-trained)} & \textbf{0.479} & \textbf{0.475} & \textbf{0.476} \\
\bottomrule
\end{tabular}
\end{table}

The overall reward rose from approximately 0.50 to 0.75.
The largest gain came from analytical sufficiency ($r_{\text{rm}}$: 0.30$\to$0.65), confirming that RL primarily improved the model's ability to organize evidence and reason from it.
Constraint dimensions remained stable: presentation stayed above 0.95, citation traceability held at 0.90--0.95, and the audit score rose from 0.60 to 0.73.
On DeepResearch Bench, the RL-trained model scores 0.476 overall, outperforming the base model (0.127), Tongyi DeepResearch (0.277), and Claude~3.7 (0.422), with gains appearing consistently across both finance and non-finance subsets.

This study closes the topology progression by showing that the basic swarm split still holds under long-horizon, tool-intensive RL with complex multi-dimensional rewards: the EnvService caching layer stabilizes client-side environmental noise, the two-stage judge design provides low-variance reward signals suitable for GRPO, and the server-side training loop accommodates the structured multi-turn workflow without modification.

\section{Automated Research}
\label{sec:auto_research}

The preceding experiments evaluate AgentJet as a training substrate across several swarm topologies.
This section examines the next consequence of the same separation: because clients carry the fast-changing experiment logic and servers preserve the expensive training state, the experiment-control loop itself can be delegated to autonomous agents.
Artificial intelligence research assistance for LLM reinforcement learning is increasingly important as training pipelines become more complex and experiments scale up.
On the one hand, conventional RL research typically involves only a few dozen hyperparameters, whereas LLM RL frameworks can include hundreds of core training hyperparameters because of hardware, model, and algorithmic diversity.
Groups of settings often have unknown effects until ablation studies are performed. For example, many vLLM engine arguments are suspected in the community to cause RL failures, yet little evidence exists to support or refute these claims. As a result, determining whether seemingly minor settings can cause significant deviations has become a substantial burden for human researchers.
On the other hand, most agentic training setups are bundled with sophisticated external services, such as sandboxes, search services, reward models, and application simulators. These external dependencies are orthogonal to research innovation, yet a minor failure or misconfiguration can invalidate algorithmic advances entirely.
For this reason, we introduce automated research pipelines in AgentJet as a downstream use of the swarm interface: they offload tedious but critical tasks, including hyperparameter investigation, infrastructure debugging, and systematic ablation, to autonomous agents, allowing researchers to focus on algorithmic innovation rather than engineering overhead.

The swarm architecture of AgentJet enables two complementary agent-assisted capabilities that facilitate RL academic research:
\begin{enumerate}
\item \emph{AgentJet automated task onboarding}, which enables researchers to rapidly create new RL tasks, test and deploy external training dependencies (such as LLM-as-judge reward services and AppWorld services) with limited human attention, convert agentic loops (white-box or black-box, human-in-the-loop or fully autonomous) into trainable pipelines, and migrate tasks from other frameworks.
\item \emph{AgentJet Alpha Auto Research (A3R) module}, which enables long-running autonomous research projects with parallel experiment execution, efficient GPU cluster utilization, and cost-effective automation powered by fully open-source models and coding agents.
\end{enumerate}

\subsection{Automated Task Onboarding}
\label{sec:task_onboarding}



The REPL substrate of Section~\ref{sec:repl} (which decouples the frequently revised agent and reward logic from the server-side optimizer and lets clients reattach in seconds) is what makes it practical to hand the entire task-setup loop to an AI coding agent. We refer to this workflow as \emph{automated task onboarding}: converting a natural-language task description into a running, verifiable RL training pipeline with minimal human attention.

A typical onboarding workflow proceeds as follows.
(1)~A researcher describes a training task in natural language, specifying the environment dynamics, agent objectives, reward criteria, and key training settings such as the number of GPUs to allocate.
(2)~The onboarding module spawns and monitors a coding agent that progressively reads the swarm training manuals, creates swarm client nodes, downloads or mocks training and test datasets, and establishes a reward function or LLM grader (e.g., using OpenJudge).
(3)~The correctness of the training loop is then verified in a real training cycle: the module initializes the swarm server node(s) in the background, attaches the client(s) to the training network, and starts training. Whenever a failure is encountered and a patch is applied, the client reattaches via the soft-restart path of Section~\ref{sec:repl} (seconds), or issues a force-restart to reset the full training process (minutes) when the change invalidates prior samples.
(4)~Once problems are resolved, training runs unattended until a human researcher inspects the curve or supplies additional requirements.


The onboarding module does not start from scratch.
AgentJet ships a set of swarm-utilization skills that guide and control coding agents from the broader community, such as OpenCode~\citep{opencode2025} and ClaudeCode~\citep{claudecode}, and the entire procedure can be carried out on a CPU-only laptop with a network connection to a remote swarm server.

Onboarding supports four distinct pathways for constructing RL training pipelines.
(1)~Researchers can create new RL tasks from scratch by describing a novel task in natural language (specifying environment dynamics, agent workflow, and reward criteria) and receiving a complete, runnable training client; this pathway is ideal for exploring new directions under a fail-fast philosophy.
(2)~Existing agents can be attached for RL training even when hidden inside black-box systems (e.g., ClaudeCode) or designed as passive responders (e.g., OpenClaw), making it possible to RL fine-tune production agents without rewriting their core logic.
(3)~RL tasks originally developed in other frameworks (e.g., OpenAI Gym, Gymnasium, or custom research codebases) can be ported to swarm-compatible clients for cross-algorithm comparison.
(4)~\emph{Non-Python agents} can be rerouted into the framework: agents implemented in other languages (TypeScript, Rust, Go) or deployed as external services can participate in swarm training, so organizations can train agents built on diverse technology stacks without rewriting them in Python.

\begin{figure}[htbp]
\centerline{\includegraphics[width=0.85\linewidth]{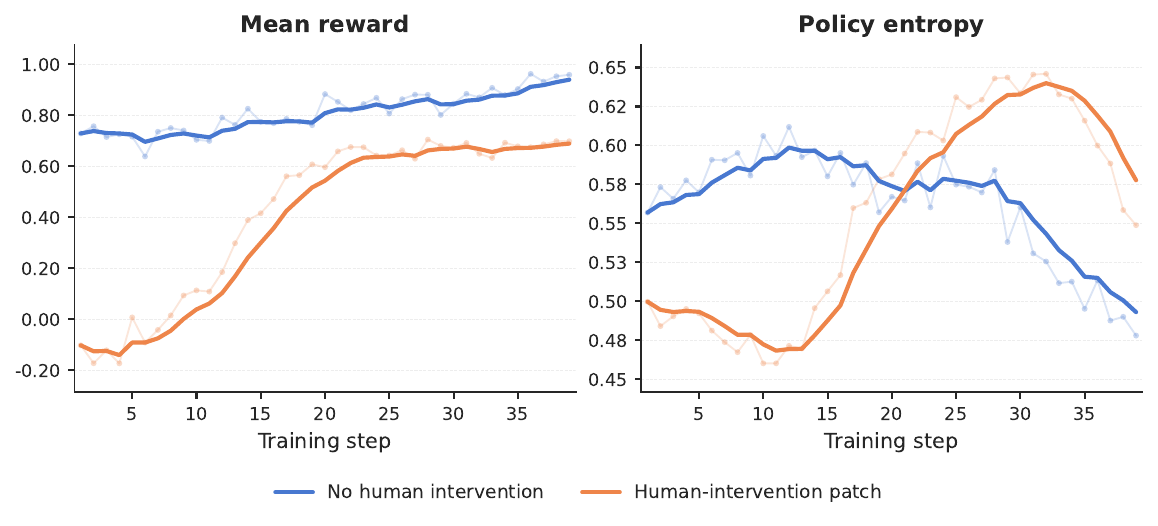}}
\caption{Example training curves for the ``Who is the Spy'' multi-agent game, a trainer constructed entirely through automated task onboarding.}
\label{fig:spy_game_rl_compare}
\end{figure}

To illustrate, we describe a case study in which a ``Who is the Spy'' multi-agent game trainer
is built entirely through natural language prompting.
The practitioner provides a single prompt specifying the game rules, the base model, the hardware budget, a request to generate mock game data,
and two training modes: cooperative (civilians share a trainable 7B model while spies use a fixed frontier model)
and adversarial (both teams use separate trainable 7B models on two swarm servers).
Figure~\ref{fig:spy_game_rl_compare} shows the resulting training dynamics for the cooperative configuration: without human intervention (red), the trainable 7B civilian policy lifts mean reward from $0.73$ to $0.96$ across $39$ GRPO steps, while policy entropy contracts (notably after step~28).

One limitation is that the onboarding module, while effective at establishing a working training loop from natural language, offers no built-in safeguard against reward hacking.
Inspecting the rollout trajectories, we found that the agents had converged on a degenerate strategy: rather than producing genuinely informative descriptions, players largely paraphrased or directly echoed prior speech from other players.
This occurs because the reward function does not penalize paraphrasing behavior.
We then applied a short patch to detect repeated player descriptions and fix the reward function; thanks to the REPL interface of Section~\ref{sec:repl}, the patched client reattached in seconds without restarting the server.
The blue curves in Figure~\ref{fig:spy_game_rl_compare} show the patched run over the same step range. Without the patch, mean reward remains artificially high from the first step ($\approx 0.73$) as agents pad their replies with copied content. With the patched reward, training instead starts from a strongly negative reward ($\approx -0.10$) and is already climbing back through zero by step~$39$ (it eventually recovers to $\approx 0.90$ over $\sim 90$ total steps); policy entropy first expands as the agent explores away from the copying attractor before contracting later in training, indicating that the agent is learning to describe the word concisely rather than mimic its peers.

\subsection{Automated Research Pipelines}

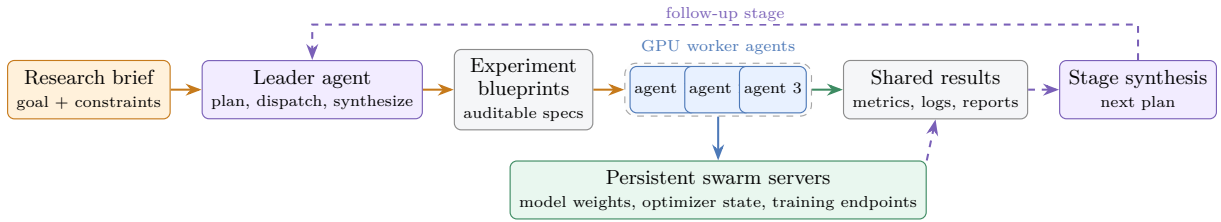
\begin{figure}[htbp]
\centering
\resizebox{\linewidth}{!}{%
\begin{tikzpicture}[node distance=0.42cm and 0.55cm]
\node[ajControl, minimum width=1.8cm] (brief) {Research brief\\\scriptsize goal + constraints};
\node[ajFeedback, minimum width=2.05cm, right=0.48cm of brief] (leader) {Leader agent\\\scriptsize plan, dispatch, synthesize};
\node[ajStore, minimum width=2.0cm, right=0.48cm of leader] (blueprints) {Experiment\\blueprints\\\scriptsize auditable specs};
\node[ajGroup, minimum width=2.9cm, minimum height=0.82cm, right=0.48cm of blueprints,
      label={[font=\scriptsize, color=ajBlueLine]above:GPU worker agents}] (workers) {};
\node[ajClient, font=\scriptsize, inner sep=2pt] at ([xshift=-0.85cm]workers.center) (w1) {agent 1};
\node[ajClient, font=\scriptsize, inner sep=2pt] at (workers.center) (w2) {agent 2};
\node[ajClient, font=\scriptsize, inner sep=2pt] at ([xshift=0.85cm]workers.center) (w3) {agent 3};
\node[ajStore, minimum width=2.0cm, right=0.48cm of workers] (results) {Shared results\\\scriptsize metrics, logs, reports};
\node[ajFeedback, minimum width=1.8cm, right=0.48cm of results] (report) {Stage synthesis\\\scriptsize next plan};

\node[ajServer, minimum width=3.1cm, below=0.68cm of workers] (swarm) {Persistent swarm servers\\\scriptsize model weights, optimizer state, training endpoints};

\draw[ajControlArrow] (brief) -- (leader);
\draw[ajControlArrow] (leader) -- (blueprints);
\draw[ajControlArrow] (blueprints) -- (workers);
\draw[ajTrainArrow] (workers) -- (results);
\draw[ajFeedbackArrow] (results) -- (report);
\draw[ajDataArrow] (workers.south) -- (swarm.north);
\draw[ajFeedbackArrow] (swarm.north east) -- (results.south);
\coordinate (fbR) at ($(report.north)+(0,0.55)$);
\coordinate (fbL) at ($(leader.north)+(0,0.55)$);
\draw[ajFeedbackArrow] (report.north) -- (fbR) -- (fbL) -- (leader.north);
\node[font=\scriptsize, color=ajPurpleLine, fill=white, inner sep=1pt]
  at ($(fbL)!0.5!(fbR)+(0,0.16)$) {follow-up stage};
\end{tikzpicture}%
}
\caption{AgentJet Alpha Auto Research (A3R) workflow over AgentJet. Client-side leader and worker agents manage experiment control, while persistent swarm servers maintain model state and training endpoints across stages and worker restarts.}
\label{fig:auto_research_architecture}
\end{figure}

The automated task onboarding module is specialized for engineering agent-loop innovations and establishing task baselines.
It is therefore a prerequisite for automation, but not a complete solution for in-depth reinforcement learning research.
Long-running RL studies add two requirements beyond task construction:
(1)~an individual experiment often lasts more than 10 hours, and sometimes several days, so the research agent must monitor, recover, and summarize long-running jobs efficiently;
(2)~ablation studies often require multiple experiments in parallel across GPU servers or cluster APIs, so the research agent must coordinate a hierarchical multi-agent structure across machines.

Within the AgentJet framework, we present the \textbf{AgentJet Alpha Auto Research (A3R)} module to meet these requirements: it converts research ideas into reliable experimental results while using multiple GPU servers to explore more configurations simultaneously.
Crucially, A3R exercises the same serving-layer separation that underpins the rest of AgentJet: the persistent swarm servers still own the model weights, optimizer state, and training endpoint, while the experiment-control logic (hypothesis formation, blueprint dispatch, monitoring, recovery, and synthesis) runs entirely on the client side.
This setup directly tests whether the serving-layer training interface can support multi-day RL experiment campaigns without placing experiment-control logic inside the trainer.

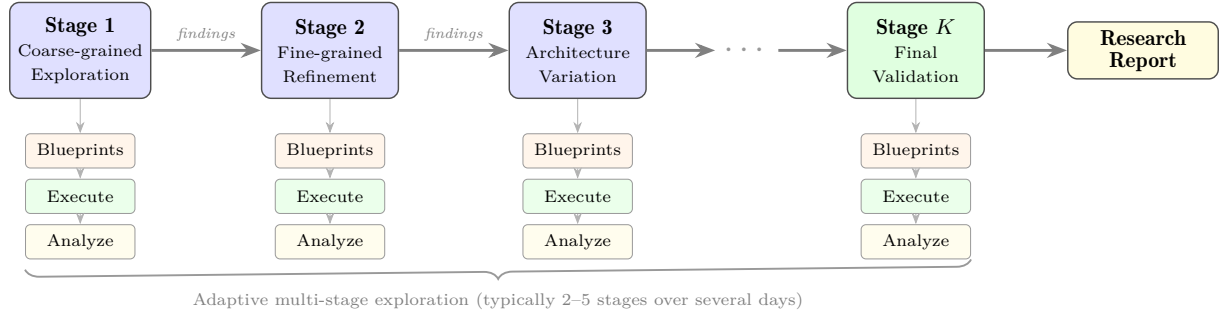
\begin{figure*}[htbp]
    \centering
    \resizebox{\textwidth}{!}{%
    \begin{tikzpicture}[
        >=Stealth,
        node distance=0.5cm,
        stage/.style={rectangle, rounded corners=4pt, draw=black!70, fill=blue!8, minimum width=2.0cm, minimum height=1.4cm, font=\small, align=center, line width=0.6pt},
        action/.style={rectangle, rounded corners=2pt, draw=black!40, fill=white, minimum width=1.6cm, minimum height=0.5cm, font=\tiny, align=center},
        arr/.style={->, very thick, color=black!50},
        looparr/.style={->, thick, color=purple!50, dashed},
    ]

    \node[stage, fill=blue!12] (s1) {
        \textbf{Stage 1}\\
        {\scriptsize Coarse-grained}\\
        {\scriptsize Exploration}
    };
    \node[stage, fill=blue!12] (s2) [right=1.6cm of s1] {
        \textbf{Stage 2}\\
        {\scriptsize Fine-grained}\\
        {\scriptsize Refinement}
    };
    \node[stage, fill=blue!12] (s3) [right=1.6cm of s2] {
        \textbf{Stage 3}\\
        {\scriptsize Architecture}\\
        {\scriptsize Variation}
    };
    \node[font=\Large, color=black!40] (dots) [right=1.0cm of s3] {$\cdots$};
    \node[stage, fill=green!12] (sf) [right=1.0cm of dots] {
        \textbf{Stage $K$}\\
        {\scriptsize Final}\\
        {\scriptsize Validation}
    };

    \node[rectangle, rounded corners=4pt, draw=black!70, fill=yellow!15, minimum width=2.2cm, minimum height=0.8cm, font=\small, align=center, line width=0.6pt] (final) [right=1.2cm of sf] {\textbf{Research}\\[-1pt]\textbf{Report}};

    \draw[arr] (s1) -- (s2) node[midway, above, font=\tiny\itshape, color=black!50] {findings};
    \draw[arr] (s2) -- (s3) node[midway, above, font=\tiny\itshape, color=black!50] {findings};
    \draw[arr] (s3) -- (dots);
    \draw[arr] (dots) -- (sf);
    \draw[arr] (sf) -- (final);

    \foreach \s/\sname in {s1,s2,s3,sf} {
        \node[action, fill=orange!8] (bp\sname) [below=0.5cm of \s] {\scriptsize Blueprints};
        \node[action, fill=green!8] (ex\sname) [below=0.15cm of bp\sname] {\scriptsize Execute};
        \node[action, fill=yellow!8] (an\sname) [below=0.15cm of ex\sname] {\scriptsize Analyze};
        \draw[->, thin, color=black!30] (\s.south) -- (bp\sname.north);
        \draw[->, thin, color=black!30] (bp\sname.south) -- (ex\sname.north);
        \draw[->, thin, color=black!30] (ex\sname.south) -- (an\sname.north);
    }

    \draw[decorate, decoration={brace, amplitude=6pt, mirror}, thick, color=black!40]
        (ans1.south west) ++(0,-0.15) -- (ansf.south east |- ans1.south west) ++(0,-0.15)
        node[midway, below=8pt, font=\scriptsize, color=black!50, align=center] {Adaptive multi-stage exploration (typically 2--5 stages over several days)};

    \end{tikzpicture}%
    }
    \caption{Multi-stage research loop. Each stage consists of blueprint generation, parallel experiment execution on GPU clusters, and result analysis. Findings from each stage inform the design of the next, enabling adaptive convergence toward the research objective.}
    \label{fig:auto_research_loop}
\end{figure*}

A3R employs two types of agents that work collaboratively.
(1) The \textbf{leader} agent plays the role of a chief scientist that forms hypotheses, designs experiments, dispatches work, collects results, and then draws conclusions or designs the next stage of experiments. There is no hardware requirement for the leader agent.
(2) The \textbf{worker} agent runs a specific experiment according to the requirements of the leader agent. Worker agents must run on GPU servers.

A3R can run long-running experiments in parallel across multiple distributed servers. Depending on the user choice,
A3R can use either cloud cluster APIs or regular SSH to access GPU servers and dispatch experiments. To coordinate distributed experiments across multiple devices, A3R uses \textbf{experiment blueprints} as the primary medium for communication between the leader and worker agents.
Each experiment blueprint is a document containing the following sections:

\begin{enumerate}
    \item \textbf{Purpose.} The objective, hypothesis, and controlled variables of the experiment.
    \item \textbf{Codebase.} The absolute path to the experiment code.
    \item \textbf{Virtual Environment.} The primary Python virtual environment path.
    \item \textbf{Configuration.} The path to the configuration file (usually YAML) specifying model, dataset, training algorithm, and all hyperparameters.
    \item \textbf{Command.} The shell command used to prepare and start training. This command is for reference only, as the worker agent may need to modify it according to runtime conditions.
    \item \textbf{Result Directory.} The path where the worker agent should write current results when the experiment finishes or fails in an unrecoverable way.
    \item \textbf{Time Budget.} The maximum allowed runtime, after which the experiment is terminated by force.
    \item \textbf{Additional Notes.} Any additional information relevant to the experiment, such as environment preparation steps, key configuration references, and pre-experiment predictions.
\end{enumerate}

This standardized format ensures that blueprints are both human-auditable and machine-parseable, and that any experiment is fully reproducible from its blueprint alone.

\begin{algorithm}[htbp]
\caption{AgentJet Alpha Auto Research (A3R) Pipeline}
\label{alg:a3r_pipeline}
\begin{algorithmic}[1]
\REQUIRE Research topic in natural language
\STATE The A3R module takes the research topic as input.
\STATE A leader agent is created to orchestrate the entire research project.
\STATE The leader agent decomposes the research topic and sketches a multi-stage experimental plan. At each stage, the leader agent plans a set of experiments to be carried out. For each experiment, the leader agent generates a detailed specification document, which we refer to as an \textbf{experiment blueprint}, containing key information such as the experiment purpose, codebase path, time budget, GPU requirements, and the recommended command line to start the experiment. Once the blueprints are generated, the leader agent uses server cluster APIs or regular SSH to deliver them to the participating GPU servers and initializes a number of \textbf{worker} agents on these servers equal to the number of blueprints.
\STATE On each distributed server, the \textbf{worker} agent conducts the experiment according to the blueprint assigned to it. Depending on the sophistication of the experiment, the worker agent may need to complete preparation jobs such as setting up the workspace, installing dependencies, and preparing datasets.
\STATE Once preparation is finished, the worker agent launches the training process and continuously monitors its progress. It diagnoses and attempts to recover from runtime errors (e.g., out-of-memory failures, GPU resource conflicts, dependency mismatches), enforces the time budget of the blueprint, and writes structured intermediate results (validation metrics, training curves, and logs) to a shared result directory.
\STATE Meanwhile, the leader agent polls the result directories of all dispatched experiments at regular intervals. As each experiment finishes, the leader ingests the structured results and accumulates them into the findings of the current stage, while still-running experiments continue without interruption.
\STATE Once all experiments in a stage complete, the leader agent synthesizes their results into a stage conclusion document, compares the observed outcome against the pre-registered contingency plan, and then decides whether the findings are conclusive enough to end the research project or whether further experimentation is warranted (in which case the loop returns to step~3).
\STATE In the final phase, the leader agent writes either a short experiment report or a full academic paper according to the user requirement, using the core experimental discoveries to populate comprehensive tables and figures. The leader agent then pauses and waits for the human researcher to provide further instructions.
\end{algorithmic}
\end{algorithm}

\subsubsection{Advantages of A3R}

A3R inherits much of its reliability from the existing swarm infrastructure of AgentJet, and a few additional design choices proved decisive in the case studies above.

\textbf{Parallel Experimentation.}
Most existing auto-research systems are designed around a single workstation and stall as soon as an experiment claims the local GPUs. A3R is built on top of the AgentJet swarm dispatch path: the leader agent ships experiment blueprints to as many GPU servers as the cluster offers, through either cloud cluster APIs or plain SSH.

\textbf{Effective Experiment Supervision.}
In general, agents without dedicated instructions tend to redirect terminal output to a file and poll it at fixed intervals when monitoring long-running experiments. However, this approach is ineffective for timely crash detection, potentially leading to wasted time and resources. A3R addresses this limitation by providing well-defined, empirically validated skills for leveraging tmux (a widely adopted terminal multiplexer on Linux). Consequently, when a worker agent is idle between observation intervals, unexpected crashes can immediately wake it for diagnosis and recovery, rather than waiting until the next scheduled check, thereby significantly accelerating the research process.

\textbf{Cost Efficiency.}
Non-open-source models are too expensive to be used as researcher agents in long-running experiments, limiting the scalability and accessibility of automated research.
A3R addresses this problem by reducing the difficulty of steps where agents are more likely to make mistakes. For example, the A3R module (1) provides a simpler CLI for remote server manipulation and (2) automatically guides agents back on track when they become stuck due to network or permission errors.

Benefiting from these guardrails, each shell step the agent issues is less likely to derail the run, so the per-decision capability bar drops and smaller open-weight models become viable. In practice, A3R can be driven by MiniMax-M2.7 to power both leader and worker agents, which makes A3R both low-cost and free of frontier-API dependencies.

\subsubsection{Research Case Study: Minimum Stable Batch Size for AIME Swarm RL}
\label{sec:aime_minbs}

To illustrate how the A3R pipeline transforms a one-paragraph natural-language brief into a full ablation study, this subsection walks through a representative case in which the leader agent was asked to identify the minimum stable training batch size for AIME swarm-mode GRPO on Qwen3-8B, and to determine whether the per-turn response budget (\texttt{max\_response\_length\_in\_one\_turn}, abbreviated \texttt{mr}) shifts that minimum. The original input is an instruction prompt; everything that follows, including stage decomposition, blueprint authoring, parallel dispatch, crash recovery, and final synthesis, was performed autonomously over roughly three calendar days and 19 PAI jobs. Table~\ref{tab:case_aime_minbs} reproduces, in chronological order, every artifact produced by the leader and worker agents.

\textbf{Protocol.}
The protocol sweeps \texttt{train\_batch\_size}$\in\{1,2,3,4,8,16,32,64\}$ on Qwen3-8B at \texttt{max\_response\_length\_in\_one\_turn}$=10000$, then probes the most informative subset at \texttt{mr}$\in\{8000, 12000\}$. All runs train for $60$ steps with GRPO ($\text{kl\_loss\_coef}=2{\times}10^{-3}$, \texttt{low\_var\_kl}, $\text{clip\_ratio}=0.2$, \texttt{ppo\_epochs}$=1$, mini-batch num $=1$), one swarm server with 8-GPU FSDP~\citep{zhao2023fsdp} per experiment, and validate on $30$ AIME-2024 problems with \texttt{val\_pass\_n}$=4$ rollouts.

The case study highlights three properties of A3R that are absent from single-shot prompting: (i)~each stage is \emph{conditional} on the previous one, with the leader narrowing the response-length probe only to the regime that the batch sweep flagged as informative, rather than committing to a fixed grid up front; (ii)~the worker agent absorbs the \texttt{bs}$=64$ interruption at step~30 by writing a partial record and queueing a clean rerun (PAI \texttt{dlcy0znkazok0pvo}) without leader or human intervention; and (iii)~the interrupted run is reported alongside the rerun in the final trace, preserving a complete audit trail rather than silently dropping the failed point.

\begin{table}[H]
\centering
\caption{Chronological trace of the AIME minimum-stable-batch-size case study. Each row is a self-contained artifact emitted by either the leader agent (planning, synthesis) or worker agents (execution). All runs use Qwen3-8B + GRPO with \texttt{ppo\_epoch}$=1$, mini-batch num $=1$, 8-GPU FSDP per experiment, and \texttt{total\_training\_steps}$=60$.}
\label{tab:case_aime_minbs}
\renewcommand{\arraystretch}{1.05}
\scriptsize
\begin{tabular}{@{}p{2.0cm} p{11.4cm}@{}}
\toprule
\textbf{Phase} & \textbf{Content} \\
\midrule
Researcher input & \emph{``Find the minimum \texttt{batch\_size} (range $1$--$64$) that still trains the AIME agent efficiently, and determine whether \texttt{max\_response\_length\_in\_one\_turn} influences that value.''} Constraints: codebase \texttt{agentjet\_codebase}, tutorial \texttt{tutorial/opencode\_build\_aime}, model \texttt{Qwen3-8B}, $\le 60$ training steps, 8 GPUs/exp, \texttt{ppo\_epoch}$=1$, mini-batch num $=1$. \\
\midrule
Leader plan & Three-stage adaptive design. \emph{Stage~1}: coarse low-end sweep \texttt{bs}$\in\{1,2,4,8\}$ at \texttt{mr}$=10000$ to bracket the lower frontier. \emph{Stage~2}: extend upward (\texttt{bs}$\in\{3,16,32,64\}$) at \texttt{mr}$=10000$ to find the upper plateau. \emph{Stage~3}: response-length probes (\texttt{mr}$\in\{8000,12000\}$) on the most informative batch sizes. Pre-registered contingency table maps frontier patterns (monotonic / saturating / non-monotonic) to next actions. \\
\midrule
Stage~1 blueprints \& execution & Four blueprints (\texttt{bs1\_mr10000}, \texttt{bs2}, \texttt{bs4}, \texttt{bs8}) at \texttt{mr}$=10000$, dispatched as PAI jobs. All four reached step~60 without infrastructure faults. \\
\midrule
Stage~1 results &
{\scriptsize
\begin{tabular}{@{}lccccc@{}}
\toprule
bs & init p@1 & final p@1 & init p@2 & final p@2 & mean reward \\
\midrule
1 & 48.33 & 36.67 & 53.33 & 53.33 & 0.3667 \\
2 & 41.67 & 41.67 & 50.00 & 50.00 & 0.4167 \\
4 & 38.33 & 45.00 & 50.00 & 60.00 & 0.4500 \\
\textbf{8} & 41.67 & \textbf{51.67} & 53.33 & 60.00 & \textbf{0.5167} \\
\bottomrule
\end{tabular}}\newline
Stability std (pass@1 across steps): \texttt{bs}$=1$: $4.47$\% (unstable); $2$: $2.59$\%; $4$: $2.44$\%; $8$: $4.18$\% (stable). \\
\midrule
Stage~1 finding & \texttt{bs}$=1$ falls below its zero-shot baseline and shows the highest step-to-step std; \texttt{bs}$\in\{2,4\}$ are stable but underperform \texttt{bs}$=8$ on the final frontier. The lower edge of the curve is established, but the upper plateau is not yet reached, so the next stage must extend toward larger batches. \\
\midrule
Stage~2 blueprints \& execution & Four blueprints (\texttt{bs}$\in\{3,16,32,64\}$ at \texttt{mr}$=10000$) dispatched 2026-04-29 19:04 CST as PAI jobs \texttt{dlc1w6i0i2ojuhax}, \texttt{dlcra32iv1d7gaae}, \texttt{dlc4d4t1mu1p0fjf}, \texttt{dlc7yzj0qe2uty4w}. The first \texttt{bs}$=64$ run was interrupted at step~30; the worker agent wrote a partial-results record, and the leader queued a clean rerun (PAI \texttt{dlcy0znkazok0pvo}) without human intervention. \\
\midrule
Stage~2 results &
{\scriptsize
\begin{tabular}{@{}lccc@{}}
\toprule
bs & final p@1 & final p@2 & mean reward \\
\midrule
3 & 43.33 & 56.67 & 0.4333 \\
\textbf{16} & \textbf{53.33} & \textbf{73.33} & \textbf{0.5333} \\
32 & 51.67 & 63.33 & 0.4997 \\
64 (clean rerun) & 50.00 & 60.00 & 0.5000 \\
\bottomrule
\end{tabular}}\newline
The interrupted \texttt{bs}$=64$ run is reported alongside the clean rerun rather than silently dropped. \\
\midrule
Stage~2 finding & \texttt{bs}$=16$ is the new frontier; \texttt{bs}$=32$ and the clean \texttt{bs}$=64$ rerun both plateau at around $50.00$\% pass@1 / $0.5000$ mean reward, slightly trailing \texttt{bs}$=16$. Increasing batch beyond $16$ does not improve performance at \texttt{mr}$=10000$. The minimum stable high-performing batch size at this response budget is \texttt{bs}$=16$. \\
\midrule
Stage~3 blueprints \& execution & Two \texttt{mr}$=8000$ blueprints (\texttt{bs}$\in\{8,16\}$) and five \texttt{mr}$=12000$ blueprints (\texttt{bs}$\in\{1,2,4,8,16\}$) dispatched as seven parallel PAI jobs. All seven reached step~60 without infrastructure faults. \\
\midrule
Stage~3 results &
{\scriptsize
\begin{tabular}{@{}llccc@{}}
\toprule
mr & bs & final p@1 & final p@2 & mean reward \\
\midrule
8000 & 8  & 38.33 & 46.67 & 0.3833 \\
8000 & 16 & 35.00 & 40.00 & 0.3500 \\
12000 & 1  & 48.33 & 63.33 & 0.4833 \\
12000 & 2  & 48.33 & 60.00 & 0.4833 \\
12000 & 4  & 56.67 & 63.33 & 0.5667 \\
12000 & 8  & 56.67 & 66.67 & 0.5667 \\
\textbf{12000} & \textbf{16} & \textbf{60.00} & \textbf{73.33} & \textbf{0.6000} \\
\bottomrule
\end{tabular}} \\
\midrule
Stage~3 finding & \texttt{mr}$=8000$ is non-competitive ($-13$ to $-18$ pts pass@1 vs. \texttt{mr}$=10000$). \texttt{mr}$=12000$ lifts the entire family; \texttt{bs16\_mr12000} is the strongest configuration overall, and the efficient lower bound under \texttt{mr}$=12000$ shifts up to \texttt{bs}$=4$ (which ties \texttt{bs}$=8$ on pass@1 and mean reward). The leader closes the loop. \\
\midrule
Final recommendation & \emph{Cheapest stable: \texttt{bs}$=4$, \texttt{mr}$=12000$. Overall best: \texttt{bs}$=16$, \texttt{mr}$=12000$.} Versus the \texttt{bs}$=8$ / \texttt{mr}$=10000$ stage-1 leader, the recommended configuration improves pass@1 by $+8.3$ pts ($51.67\!\to\!60.00$) and pass@2 by $+13.3$ pts ($60.00\!\to\!73.33$). Larger batches (\texttt{bs}$\geq 32$) are explicitly not recommended: they consume more compute without moving the frontier. \\
\bottomrule
\end{tabular}
\end{table}

\textbf{Findings.}
Three conclusions follow from the chronological trace in
Table~\ref{tab:case_aime_minbs}.
\textbf{(i)~The minimum stable high-performing batch size at \texttt{mr}$=10000$
is \texttt{bs}$=16$.} Below this point pass@1 and mean reward both degrade
monotonically (\texttt{bs}$=1$ falls below its zero-shot baseline and exhibits
$4.47$\% step-to-step std), and very small batches fail to integrate enough
GRPO group statistics to produce a stable advantage signal.
\textbf{(ii)~Increasing batch beyond \texttt{16} does not move the
\texttt{mr}$=10000$ frontier:} \texttt{bs}$=32$ and the clean \texttt{bs}$=64$
rerun both finish at $50.00$\% pass@1 / $0.5000$ mean reward, slightly trailing
\texttt{bs}$=16$. Extra GPUs are therefore better spent on the response budget
than on a wider batch, a recommendation A3R surfaces directly in its final report.
\textbf{(iii)~Response length dominates batch size for this workload.}
Shrinking \texttt{mr} to $8000$ collapses both \texttt{bs}$=8$ and \texttt{bs}$=16$
($-13$ to $-18$ pts pass@1), while extending \texttt{mr} to $12000$ lifts the
entire family: \texttt{bs16\_mr12000} is the strongest configuration at
$60.00$\% pass@1, $73.33$\% pass@2, and $0.6000$ mean reward, and the efficient
lower bound shifts up to \texttt{bs}$=4$ (which ties \texttt{bs}$=8$ on pass@1
and mean reward at \texttt{mr}$=12000$). \texttt{mr}$=8000$ is non-competitive
in this setup.

\section{Conclusion}
\label{sec:conclusion}

We presented \textbf{AgentJet}, a distributed swarm framework for agentic reinforcement learning that separates the model-optimization plane from the agent-execution plane through an OpenAI-compatible serving-optimization protocol. Because training-side state (model weights, optimizer state, context tracking, and sample pools) stays on GPU-side servers while request-based agents run unchanged, AgentJet makes the server--client topology a configurable design dimension of agentic RL. Reconfiguring the same interface allows researchers to move among multi-model, multi-task, fault-tolerant, and REPL-style training setups without redesigning the trainer.

To make this decoupled interface trainable and efficient, AgentJet adds per-episode context tracking, sample-pool--driven batching, and timeline merging; on an AppWorld workload, the conservative default of timeline merging cuts actor-update wall-clock time by 6.25$\times$ on average while preserving the observed rollout behavior, whereas more aggressive settings trade that rollout fidelity for additional speed. Across the topology progression evaluated in this paper, the same interface supports shared-parameter training, non-shared-parameter heterogeneous multi-model training, mixed-task cocktail training over isolated runtimes, and regular multi-turn training with framework-agnostic clients.

Finally, the same infrastructure broadens automation: an \emph{automated task onboarding} module converts agentic loops into trainable pipelines with minimal setup, and the Alpha Auto Research (A3R) module drives multi-day RL campaigns with parallel GPU utilization and adaptive ablation. Across our studies (four agent-loop implementations, shared- and non-shared-parameter Werewolves training, pipeline-internalized academic translation, mixed-task RL over isolated AppWorld and AIME runtimes, long-horizon financial research training, and multi-day AutoResearch campaigns), AgentJet provides an agent-oriented I/O layer for RL in which execution-side logic remains configurable while optimization-side state remains persistent.

\bibliographystyle{colm2025_conference}
\bibliography{references}

\end{document}